\documentclass{article}

\usepackage[preprint]{neurips_2026}

\usepackage[utf8]{inputenc} 
\usepackage[T1]{fontenc}    
\usepackage[hidelinks]{hyperref}       
\usepackage{url}            
\usepackage{booktabs}       
\usepackage{amsfonts}       
\usepackage{nicefrac}       
\usepackage{microtype}      

\usepackage{amsmath}
\usepackage{amssymb}
\usepackage{mathtools}
\usepackage{amsthm}
\usepackage[table,xcdraw]{xcolor}         
\usepackage{adjustbox}
\usepackage{multirow}
\usepackage{boldline}
\usepackage{booktabs}
\usepackage{enumitem}
\usepackage{algorithm}
\usepackage{algorithmic}
\usepackage{arydshln}
\usepackage{ifsym}

\usepackage{array}
\usepackage{makecell}

\usepackage{fontawesome5}

\usepackage{graphicx}
\newcommand{\huggingface}{
  \raisebox{-0.2em}{\includegraphics[height=1em]{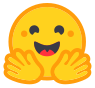}}
}

\newcommand{\normwear}{
  \raisebox{-0.15em}{\includegraphics[height=1.1em]{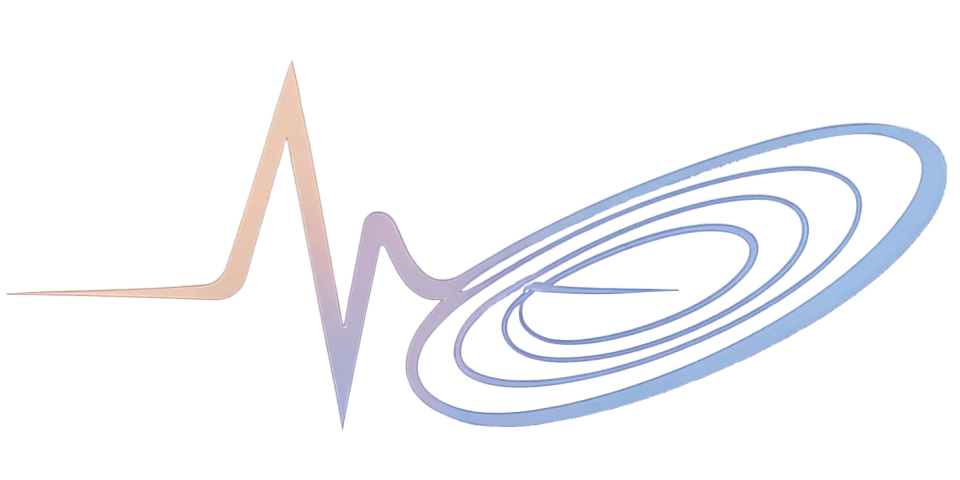}}
}

\usepackage{pifont} 

\setlist[itemize]{topsep=-0.18cm}

\newcommand{\edit}[1]{\textcolor{black}{#1}}

\title{\normwear Toward World Modeling of Physiological Signals with Chaos-Theoretic Balancing and Latent Dynamics}

\author{
Yunfei Luo$^{1}$\footnote[2]{} \qquad
Xi Chen$^{1}$ \qquad
Yuliang Chen$^{1,2}$ \qquad
Lanshuang Zhang$^{1}$
\\
\textbf{
Md Mofijul Islam$^{3}$\thanks{Work does not relate to position at Amazon.} \quad
Siwei Zhao$^{4}$ \quad
Peter Kotanko$^{5,6}$ \quad
Subhasis Dasgupta$^{1}$
} \\
\textbf{
Andrew Campbell$^{2}$ \qquad\;\;
Rakesh Malhotra$^{1}$ \qquad\;\;
Tauhidur Rahman$^{1}$\thanks{Corresponding authors: \texttt{\{yul268, trahman\}@ucsd.edu}}
} 
\vspace{2mm} \\
$^{1}$University of California San Diego \quad
$^{2}$Dartmouth College \quad
$^{3}$Amazon Web Services \\
$^{4}$Sanderling Renal Services 
$^{5}$Renal Research Institute 
$^{6}$Icahn School of Medicine at Mount Sinai 
\vspace{2mm} \\
\huggingface \textbf{\textsc{NormWear} Collection:} \href{https://huggingface.co/collections/mosaic-laboratory/normwear}{\texttt{mosaic-laboratory/normwear}}
\vspace{-3mm}
}

\begin{document}

\maketitle

\begin{abstract}
Physiological time series signals reflect complex, multi-scale dynamical processes of the human body. Existing modeling studies focus on static tasks such as classification, event forecasting, or short-horizon next step prediction, while long-horizon signal-level forecasting and predictive nature of physiological signals remain underexplored.
We introduce \textbf{NormWear-2}, a world model that encodes both multivariate physiological signals and clinical intervention variables into a shared latent space and models their joint temporal evolution as a dynamical system. Our approach combines inference from prior pre-trained knowledge (\emph{intuition}) with instant non-parametric latent state transition adaptation (\emph{insight}), enabling coherent forecasting across multiple temporal scales, conditioned on heterogeneous clinical interventions.
During the pretraining phase, we find that chaos-theoretic balancing of dynamical regime diversity yields more robust representations, with a smaller balanced corpus outperforming one twice its size and capturing bifurcation regimes. 
We evaluate the world model performance across diverse real-world physiological datasets spanning heterogeneous temporal resolutions and intervention regimes, covering daily life, point-of-care, and clinical settings, including fitness planning, hemodialysis, diabetes management, and surgical monitoring. These evaluation datasets comprise records from 8,026 subjects, spanning study durations from 3.2 hours for high-resolution signal data to 2.3 years for longitudinal clinical biomarker tracking. NormWear-2 achieves the best overall forecasting performance across time, frequency, and latent representation domains, with significant improvements over state-of-the-art time series foundation models, while maintaining competitive downstream representation quality, providing a step toward general-purpose world models for physiological signals.
\end{abstract}

\begin{figure*}[t!]
\centering
\includegraphics[width=1.0\textwidth]{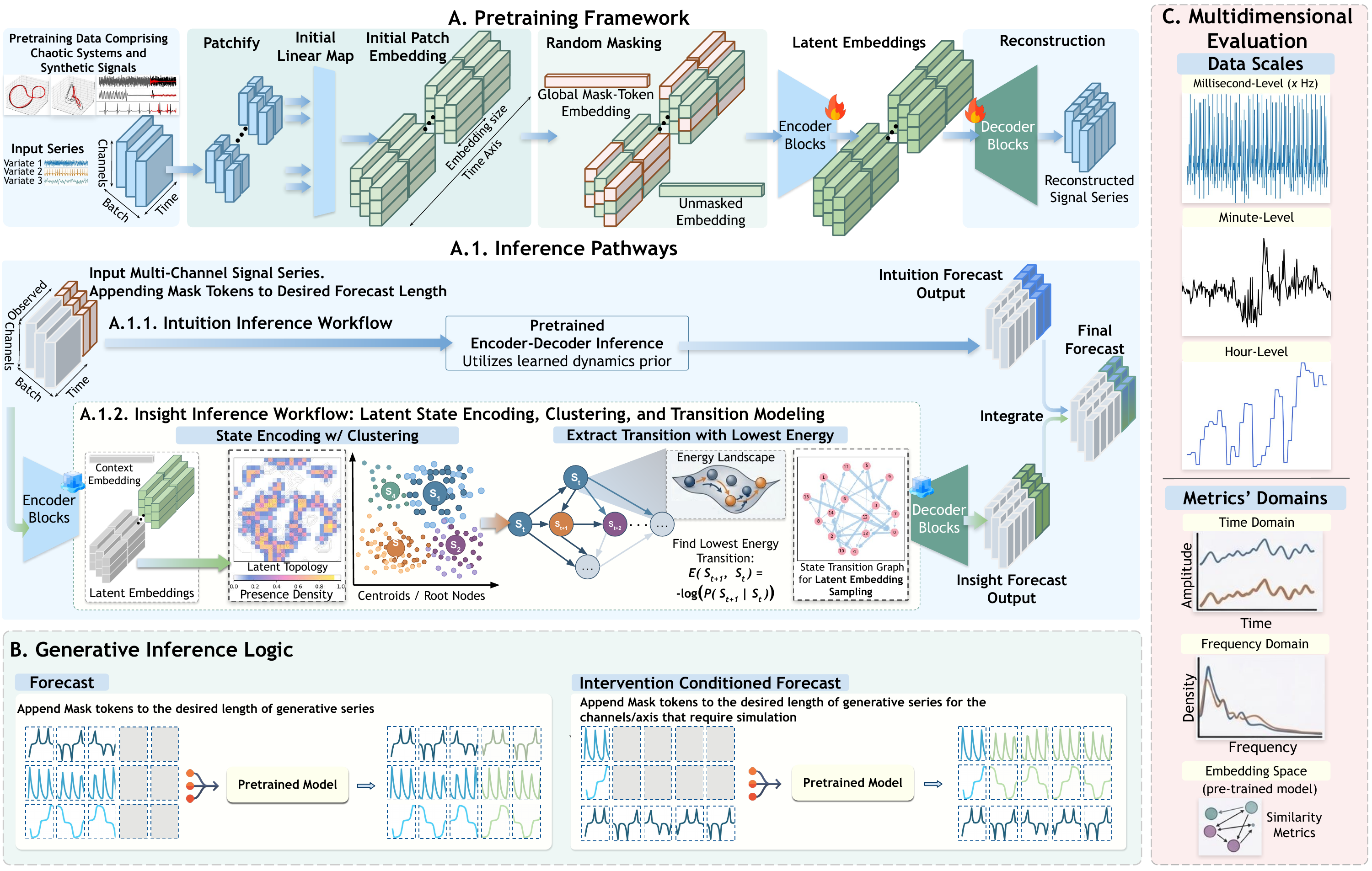}
\vspace{-6mm}
\caption{\textbf{Methodology.} 
(A) Overview of the modeling workflow from the input signals to pretraining and forecasting output. 
(A.1.) Proposed intuition-insight inference pathways. 
(B) Demonstration of the generative prediction logic after the standard mask-and-reconstruction pretraining. 
(C) Multidimensional evaluation across multiple temporal resolution and performance metrics. 
}
\label{fig:overview}
\vspace{-4mm}
\end{figure*}

\vspace{-2mm}
\section{Introduction} \label{sec:intro}
\vspace{-2mm}

Physiological signals provide a continuous and non-invasive window into the internal dynamics of the human body. Modalities such as electroencephalography (EEG), electrocardiography (ECG), and photoplethysmography (PPG) encode rich temporal patterns spanning multiple scales, from milliseconds to hours. These signals are inherently generated by complex dynamical systems, yet most machine learning approaches treat them as static inputs for downstream tasks such as classification, regression, or anomaly detection \citep{papagei,hire-mae,normwear}. As a result, they lack the ability to reason about future trajectories, simulate alternative scenarios, or adapt to changing physiological states. 

More broadly, real-world multivariate time series often arise from systems exhibiting diverse dynamical behaviors, ranging from quasi-periodic and limit-cycle patterns to weakly or strongly chaotic regimes \citep{nld-and-chaos,autoformer,batterylife,panda}. Differences in observed time series frequently reflect variations in predictability, sensitivity to initial conditions, and structural complexity rather than fundamentally distinct generative mechanisms. Capturing such nonlinear dynamics is critical for learning representations that generalize across systems and domains. 
Recent advances in representation learning have improved performance in general time-series modeling \citep{chronos2,sundial,panda,chaosbenchmark}. However, these works did not explicitly investigate in the distribution of underlying dynamical regimes while pretrain the base model. This can not only diminish our understanding toward the behaviors of model learning and inference, but also limit the robustness and transferability of learned representations. Furthermore, the feasibility of these approaches on physiological signals and digital healthcare applications involving intervention variables remain underexplored. 

To address these challenges, we propose NormWear-2, a framework for latent world modeling of physiological signals as briefly demonstrated in Figure \ref{fig:overview}. The model encodes multimodal time series into a shared latent space where temporal evolution is modeled as a dynamical system, combining two complementary mechanisms: (1) \textit{intuition}, capturing prior knowledge of physiological dynamics learned from large-scale pretraining, and (2) \textit{insight}, refining latent state transitions through instant non-parametric unsupervised adaptation. This dual mechanism enables coherent multi-step forecasting and generative modeling while adapting to new observations and contexts. 
During the modeling exploration, we also discover that the effectiveness of latent world modeling depends on the diversity of dynamical regimes present in pretraining data. Leveraging chaos-inspired metrics such as Lyapunov exponents, detrended fluctuation analysis, and persistent entropy \citep{lyapunov,dfa,persistent-entropy}, we quantify dynamical properties of time series and construct pretraining corpora with balanced coverage across different dynamical behaviors. Empirically, models pretrained on dynamically balanced datasets produce more robust and transferable representations, supporting downstream predictive and generative tasks. 
Finally, we introduce a multidimensional evaluation framework that assesses forecasting quality across temporal, frequency, and latent representation domains. Experiments on a range of real-world physiological datasets demonstrate that NormWear-2 achieves accurate and consistent multi-scale predictions, highlighting the potential of latent dynamical modeling for predictive world modeling in physiological signal analysis.

\vspace{-1mm}
\section{Related Work}
\vspace{-2mm}
%
In physiological domains, most approaches, ranging from task-specific architectures \citep{convtran,cbramod,ecgfm} to pretrained representations \citep{normwear}, emphasize supervised or self-supervised representation learning for downstream tasks, rather than explicit modeling of temporal dynamics for forecasting or simulation. This limits their ability to support predictive reasoning under interventions.
%
Dynamical systems approaches, including state-space and latent variable models, provide a principled framework for modeling temporal evolution \citep{ieee_latent_cluster}. More recently, world models and latent predictive learning have emphasized learning structured representations for forecasting \citep{maes2026leworldmodel,nam2026causal}. 
%
From a modeling perspective, existing world modeling approaches can be broadly categorized into generative-based methods \citep{chronos2,panda,sundial} and joint-embedding predictive architectures (JEPA) \citep{maes2026leworldmodel}. Generative approaches explicitly model the conditional distribution of future trajectories. Recent time series foundation models \citep{timesfm,lagllama,moirai,autoformer} provide a strong backbone for this paradigm,
while model inference behavior across varied temporal resolutions, and the feasibility of applying such models to physiological signal forecasting in scenarios involving intervention variables, also referred to as world modeling or physiological simulation, remain largely underexplored. 
In contrast, JEPA-based approaches \citep{fox2025physiojepa} primarily focus on learning predictive representations and have been explored in physiological domains mainly for improving downstream inference tasks, rather than signal level forecasting accuracy. 
As a result, this work fills these research gaps and proposes a well-rounded, end-to-end methodology for world modeling in physiology and digital health domain. 


\vspace{-1mm}
\section{Method} \label{sec:method}
\vspace{-2mm}

\subsection{Problem Setup}
\vspace{-2mm}

World modeling, according to \citet{maes2026leworldmodel,zhiting-world-model-tutorial}, refers to a model that maintains an internal understanding of the world and is able to forecast future trajectories given current and past observations. In the field of physiological signal modeling, such a paradigm remains under-explored, as discussed in the introduction and related work. In the context of digital healthcare, we define the world model as a dynamical system, which is commonly formulated as 
$\mathbf{x}_{t+1} = f_\theta(\mathbf{x}_{\leq t}, \mathbf{u}_{\leq t})$,
where $\mathbf{x}_t$ denotes the system state and $\mathbf{u}_t$ represents the action or intervention at time $t$. 

In the context of physiological signals, we instantiate this formulation as a \emph{multivariate time series system}, where both physiological measurements and interventions evolve over time. This naturally leads to a conditional forecasting formulation: $p_\theta(\mathbf{x}_{t+1:t+H} \mid \mathbf{x}_{\leq t}, \mathbf{u}_{\leq t+H})$,
where $H$ denotes the prediction horizon. 
Under this view, world modeling in healthcare becomes a channel-conditioned multivariate time series forecasting problem, because in many clinical scenarios, interventions are inherently temporal and can be directly aligned with physiological signals. For instance, in surgical settings, machine-controlled parameters such as respiration and anesthesia delivery are continuous time series that evolve synchronously with physiological variables. Similar patterns arise in dialysis and other critical care scenarios. 

For event-based interventions, such as insulin administration or meal intake in diabetes management, we convert them into step-function representations series, where non-zero values span the duration of the event. More abstract lifestyle factors, such as physical activity, can be encoded via a pretrained language model \citep{bio_clinicalBERT} into a finite semantic space during data preprocessing, and then transformed in the same way as the event-based interventions. These strategies enable a unified temporal representation of heterogeneous interventions across diverse healthcare scenarios.

\begin{table*}[t!]
\vspace{-7mm}
\centering
\small
\caption{Datasets for evaluation. The time duration in this table referring to the average time spanning of the data for each subject. Additional information of each dataset is provided in Appendix \ref{app:datasets}.}
\renewcommand{\arraystretch}{1.0}
\resizebox{\textwidth}{!}{
\begin{tabular}{
    l
    c
    c
    c
    l
    l
}
\hlineB{2}
\textbf{Dataset} & 
\textbf{\# Subjects} & 
\textbf{Time Duration} & 
\textbf{Scale \& Scope} & 
\textbf{Physiological Signals} & 
\textbf{Intervention Variables} \\
\hline

\makecell[l]{VitalDB \\ \citep{vitaldb}}
& 6,388 
& 3.2 Hours 
& \makecell[c]{Millisecond, \\ Wearable and Medical \\ Device Sensing} 
& \makecell[l]{ECG, PPG, EEG, \\ Respiration} 
& \makecell[l]{
Remifentanil Target Concentration, \\ 
Fraction of Inspired Oxygen, \\
Fresh Gas Flow Rate (L/min), \\
Inspiratory Time (s), Respiratory Rate,  \\
Tidal Volume (mL)
} \\
\hline
\makecell[l]{PMData \\ \citep{pmdata}} 
& 16 
& 5 Months 
& \makecell[c]{Minute, \\ Sport Wristband Sensing} 
& \makecell[l]{Heart Rate, Steps, \\ Distance, Calories} 
& \makecell[l]{Lifestyle Sports, e.g. Running, \\Soccer, Strength, etc.} \\
\hline
\makecell[l]{CGMacros \\ \citep{cgmacros}} 
& 45
& 10.9 days
& \makecell[c]{Minute, \\ Biofluidic Sensing} 
& \makecell[l]{Glucose, Heart Rate, \\Physical Motion} 
& \makecell[l]{Food Nutrition Composition} \\
\midrule 
\makecell[l]{Shanghai Diabetes \\ \citep{shanghai_diabete}} 
& 125
& 10.7 days
& \makecell[c]{Quarter Hour, \\ Biofluidic Sensing} 
& \makecell[l]{Glucose, Heart Rate} 
& \makecell[l]{Insulin, Hypoglycemic Agents} \\
\hline
\makecell[l]{KidneyDialysis \\ \citep{idh}} 
& 1,452 
& 2.3 Years 
& \makecell[c]{Hour, \\ Medical Device Sensing} 
& \makecell[l]{Heart Rate, \\ Blood Pressures, \\ Body Temperature} 
& \makecell[l]{Rates of Blood Flow and Dialysate \\
Flow, Dialysate Temperature, \\
Ultrafiltration Rate} \\

\hlineB{2}
\end{tabular}
}
\label{tab:dataset_summary}
\vspace{-4mm}
\end{table*}

\vspace{-2mm}
\subsection{Datasets}
\vspace{-2mm}
To study world modeling of physiological signals, we require datasets that are multivariate, longitudinal, and include explicit action/intervention variables. We leverage five real-world physiological datasets for core method evaluation: VitalDB \citep{vitaldb}, PMData \citep{pmdata}, Shanghai Diabetes \citep{shanghai_diabete}, CGMacros \citep{cgmacros}, and a clinical kidney dialysis (KidneyDialysis) dataset \citep{idh}. A summary of the core attributes of these datasets are provided in Table \ref{tab:dataset_summary}. 

Specifically, VitalDB is a perioperative monitoring dataset with physiological signals and anesthesia machine parameters for surgical patients. 
PMData is a wearable health dataset with heart rate, activity, sleep, and exercise records from daily life monitoring. 
CGMacros is a diabetes dataset with glucose monitoring, meal records, and activity information for personalized nutrition analysis.
Shanghai Diabetes is a real-world diabetes dataset with glucose records, medications, insulin, and dietary information. 
The KidneyDialysis dataset is a hemodialysis dataset with physiological signals and dialysis machine parameters during treatment sessions. 
Overall, these datasets cover representative healthcare scenarios including clinical monitoring, daily health management, point of care treatment, and clinical treatment, providing diverse physiological observations and intervention variables that make them well suited for evaluating healthcare world models.

For downstream tasks, we follow the same evaluation datasets as in \citet{normwear} and adopt the same evaluation setup. For studying data-balance-aware pretraining, we primarily use datasets from \citep{panda,normwear}. As this preliminary exploration study is closely related to dynamical system modeling, we additionally include benchmark datasets from prior works for validation, including \citet{autoformer,batterylife,chaosbenchmark}. Detailed datasets information are reported in Appendix \ref{app:datasets}.

\vspace{-2mm}
\subsection{Chaos-Theoretic Metrics for Data-Balance-Aware Pretraining} \label{sec:data-balance}
\vspace{-2mm}
We leverage a set of metrics from nonlinear dynamical systems and chaos theory to characterize and quantify different types of chaotic processes present in a collection of observed time series. These metrics include the detrended fluctuation analysis (DFA) exponent \citep{dfa}, the Lyapunov exponent (LE) \citep{lyapunov,lyapunov-thres}, and persistent entropy (PE) computed on zero- and one-dimensional homology \citep{persistent-entropy,persistent-entropy-stable}. Respectively, these metrics assess long-range autocorrelation, sensitivity to initial conditions, and the connectivity and loop complexity of the transformed topological structure of a time series.
Based on these chaos metrics, different dynamical system types can be identified through a deterministic pipeline that computes the metrics, applies unsupervised clustering, and assigns cluster labels using fixed, literature-established thresholds \citep{dfa,lyapunov-thres,persistent-entropy}.
To examine dataset balance with respect to chaotic behavior, we perform K-means clustering on the computed metrics for each time series sample. The optimal number of clusters is selected using the elbow method. Details of the clustering procedure and the interpretation of each cluster in terms of underlying dynamical systems are provided in Appendix~\ref{app:nld-cluster-process} and Appendix~\ref{app:chaotic-cluster}.
As shown in Figure~\ref{fig:balance_results} (panel A), the pretrained multivariate time-series benchmark datasets from \citet{panda} and \citet{normwear} are each dominated by a single type of chaotic pattern. In contrast, when these benchmarks are aggregated, the resulting dataset exhibits a more homogeneous distribution of chaotic types. This observation is consistent across both the statistical bar plots and the t-SNE visualizations.

\vspace{-2mm}
\subsection{Pretraining and Generative Inference} \label{sec:backbone}
\vspace{-2mm}
\textbf{Model backbone.}
We use the channel-aware mechanism for multivariate signal modeling proposed by \citet{normwear}. Detailed complexity analysis of multivariate time series modeling approaches \citep{normwear,panda,sundial} are presented in Appendix \ref{app:channel-aware-complex}. Figure \ref{fig:overview} panel A, with the same encoding backbone as proposed by \citet{normwear}, but optimized logic for initial time series patch embedding and the lightweight decoding block to better adapt the scenario for generative based multivariate time series modeling purpose. 

\textbf{Training and Inference.}
The backbone model is pretrained on the aggregated pretraining data benchmark as discussed in Section \ref{sec:data-balance}, in a masking and reconstructing manner. After the input multivariate time series being patchified, the patches are randomly replaced with a trainable unified [MASK] token representation with a fixed probability threshold pre-defined following guidance from \citet{audio-mae}. The masks applied on the input are independently sampled for each channel in the multivariate input, thus, varied masking combination are expected to be covered as more pretraining iterations progresses. 
During inference time, we focus on two types of generative tasks in this study: forecasting and simulation. For forecasting, the model predicts the future time series data given the past time series data. On the other hand, simulation task involves completing the unobserved channel conditioned on one or more given or observed channel. We refer this task as simulation because it naturally align with varied application scenarios such as health intervention and battery testing where we may have one or more controlled variables represented as separate input time series channels. The overview is presented in Figure \ref{fig:overview} (panel E).

\vspace{-2mm}
\subsection{Dynamical State Transition Modeling in Latent Space}
\vspace{-2mm}

After pretraining, the dynamical state transition modeling is performed during inference time. Given an observed context, the input is first encoded into latent representations via the pretrained encoder, which are then grouped into clusters to form discrete latent states. The optimal number of clusters is determined using the elbow rule, and empirically we observe that it scales approximately with the logarithm of the total number of patches extracted from the observed context.
To capture temporal dynamics, we estimate state transitions directly from consecutive patch pairs. Let $s_t \in \mathbb{R}^{emb\_size}$ denote the latent cluster assignment at time step $t$ where $emb\_size$ representing the embedding size of the backbone model. The empirical transition probability is
\begin{equation}
\resizebox{0.94\linewidth}{!}{$
P(s_{t+1}=j \mid s_t=i)=
\frac{\sum_t \mathbf{I}[s_t=i,\,s_{t+1}=j]}
{\sum_t \mathbf{I}[s_t=i]},
\qquad
s'_{t+1}\sim
\sum_j P(s_{t+1}=j \mid s_t=i)\,\mathcal{N}(\mu_j,\sigma_j^2),
\qquad
\mu_j,\sigma_j^2 \in \mathbb{R}^{emb\_size}
$}
\label{eq:expect_transit}
\end{equation}
where $\mathbf{I}[\cdot]$ is the indicator function, and $\mu_j$ and $\sigma_j$ denote the centroid and standard deviation of cluster $j$. Thus, forecasting is performed by first sampling the next latent state from the transition matrix and then drawing the latent representation from the corresponding Gaussian component.
When intervention or action variables are available, the state transition becomes action-conditioned. Specifically, the transition probability from state $i$ to state $j$ is no longer determined solely by the current state, but also by the applied action $a_t$. The marginal transition can be expressed as
\begin{equation}
\resizebox{0.94\linewidth}{!}{$
P(s_{t+1}=j \mid s_t=i)
=
\sum_{a \in A}
P(s_{t+1}=j \mid a, s_t=i)\,P(a \mid s_t=i),
\quad
s'_{t+1}
\sim
\sum_j
P(s_{t+1}=j \mid a_t, s_t=i)\,
\mathcal{N}(\mu_j,\sigma_j^2)
$}
\label{eq:action_conditioned_transition}
\end{equation}
where the first term describes the marginalization over possible actions $A$, and the second term represents the action-conditioned forecasting distribution. This formulation is equivalent to the decomposition of the expected transition output presented in Equation~\ref{eq:expect_transit}.
When action variables are introduced as additional multivariate input time-series channels, inference becomes more straightforward. At each time step, the latent state representation is concatenated with the corresponding action vector, forming an action-aware state embedding. The transition decomposition can then be directly obtained through straightforward Euclidean distance based neighborhood search over these joint state-action representations, and state transitions are subsequently restricted to the retrieved neighboring states, ensuring that forecasting remains consistent with the observed intervention dynamics. 
This transition model can alternatively be interpreted in an energy-based formulation, where an energy function is defined as $E(s_{t}, s_{t+1}) = -\log P(s_{t+1} | s_t)$. Under this view, forecasting corresponds to sampling low-energy transitions, which is equivalent to maximizing the likelihood in the corresponding Markov graphical model.
After generating latent representations for the desired number of future time steps, all patches are passed through the pretrained decoder to reconstruct the output time series. We refer to the direct output from the pretrained model as ``intuition'', as it depends entirely on the pretrained backbone, while the proposed dynamical transition modeling stage is termed ``insight'', since it incorporates information from the observed context during inference.

\vspace{-2mm}
\subsection{Multidimensional Evaluation}
\vspace{-2mm}
Beyond evaluating performance across multiple temporal scales, we propose a multidimensional evaluation protocol that captures diverse aspects of forecasting quality. Rather than relying solely on step-wise deviation from ground truth, our evaluation framework incorporates complementary metrics that assess structural, spectral, and representation-level alignment. Specifically, we consider three categories of metrics. First, for point-wise accuracy, we adopt Mean Absolute Error (MAE) to measure the average deviation between predicted and ground-truth sequences. Second, to account for temporal and morphological consistency, we employ Dynamic Time Warping (DTW), implemented via the differentiable SoftDTW formulation, which allows flexible alignment between sequences with temporal distortions. Third, to evaluate frequency-domain characteristics, we compute the cosine similarity (FreqCosSim) and Euclidean distance (FreqEucl) between the Fast Fourier Transform representations of predicted and ground-truth signals, capturing discrepancies in spectral components. Furthermore, to assess high-level semantic consistency, we leverage a pretrained encoder to extract latent embeddings from both predicted and ground-truth sequences. We then compute cosine similarity (LatentCosSim) and Euclidean distance (LatentEucl) in the latent space, providing a measure of abstract representation alignment beyond observable signal space. A final score is also keep track during the experiments to better compare the overall performance of different approaches. 
Each test sample is Z-normalized on the observed context, and the ground truth of unobserved part, which is to be predicted by the model, is also z-normalized based the mean and standard-deviation from the observed context. The scores are normalized to make sure all the metrics live on consistent numerical scales, thus, we have final score computed as:
\begin{equation}
\resizebox{0.94\linewidth}{!}{$
    \frac{1}{6}\cdot \left( MAE + \frac{Soft\_DTW}{pred\_length}  + (1-FreqCosSim) + \frac{FreqEucl}{0.5\cdot pred\_length} + (1-LatentCosSim) + \frac{LatentEucl}{embed\_size} \right)
$}
\label{eq:final_score}
\end{equation}
Where $pred\_length$ depends on the application scenario as specified in Table \ref{tab:dataset_lengths}, and scale it by $0.5$ representing the max number of unique frequency components. 
Together, these metrics offer a comprehensive evaluation framework that holistically reflects forecasting performance across temporal alignment, spectral fidelity, and semantic representation.

\begin{figure*}[t]
\vspace{-4mm}
\centering
\includegraphics[width=1.0\textwidth]{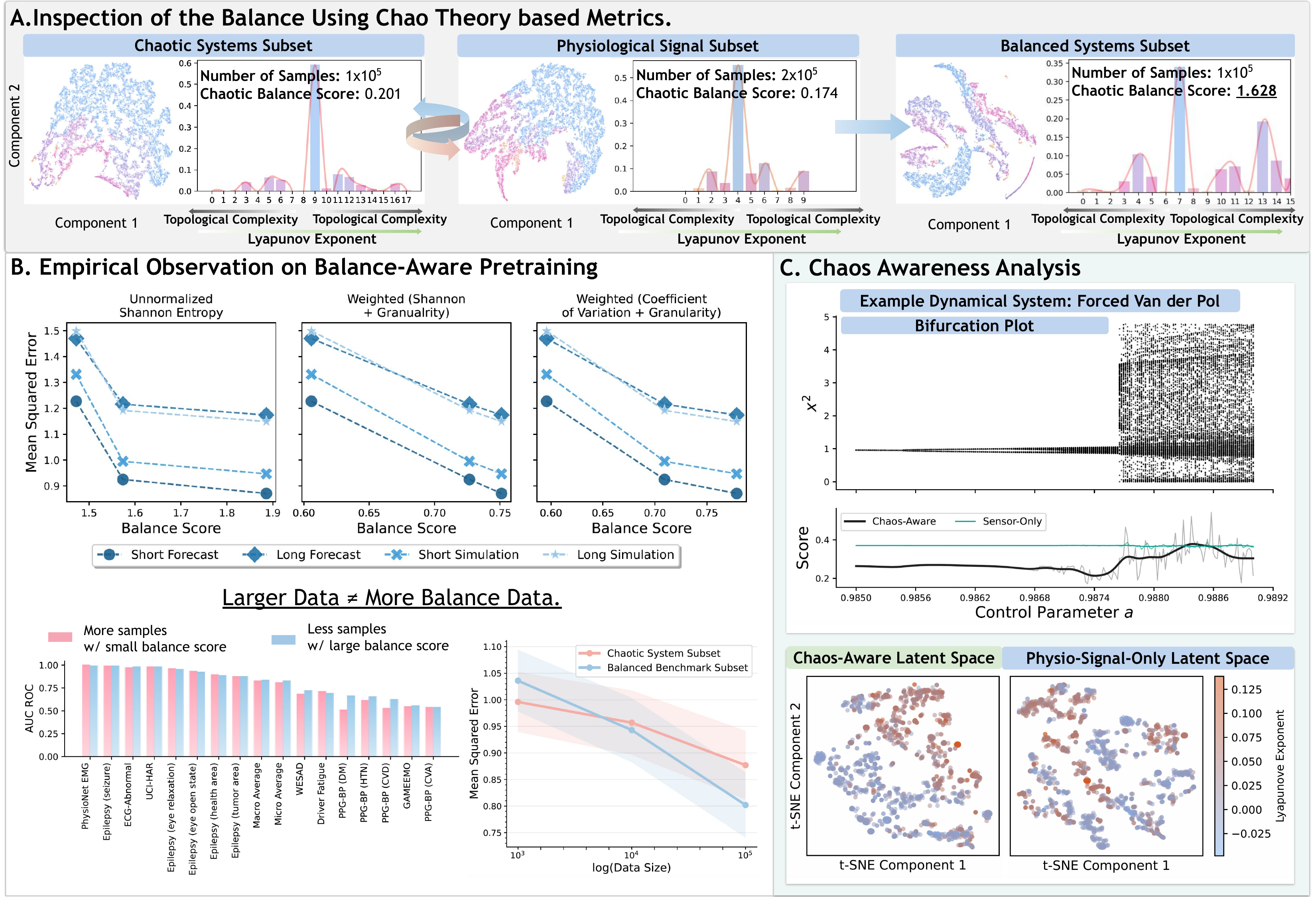}
\vspace{-7mm}
\caption{
(A) Inspection of the balance using chao theory based metrics.
(B) Balance-aware behavior: We show performance on generative tasks across models pre-trained on datasets with varied balance score. 
Details of the metrics of balance are presented in Appendix \ref{app:data-balance}. 
(C) Chaos-Aware analysis.
  }
\label{fig:balance_results}
\vspace{-4mm}
\end{figure*}

\vspace{-1mm}
\section{Experiment and Result} \label{sec:results}
\vspace{-2mm}

\begin{table*}[t!]
\centering
\small
\caption{Performance across multiple time scales under the proposed evaluation criteria.
NormWear-2 achieves the best overall performance by maintaining a consistently strong trade-off across all metrics.
}
\renewcommand{\arraystretch}{1.1}
\resizebox{0.99\textwidth}{!}{
\begin{tabular}{lccccccc}
\hlineB{2}
Model 
& MAE $\downarrow$ 
& Soft-DTW $\downarrow$ 
& FreqCos\_sim $\uparrow$ 
& FreqEucl $\downarrow$ 
& LatentCosSim $\uparrow$ 
& LatentEucl $\downarrow$ 
& \bf Final Score $\downarrow$ \\
\hline

\multicolumn{8}{c}{\textbf{Millisecond Level (VitalDB, wearable and medical device sensing)}} \\
\hline
Naive           & 0.772	& 814.595 &	0.105 &	1393.951 &	0.497 &	234.204 & 0.769 \\
Seasonal Naive  & 0.909	& 608.661 &	0.393 &	1581.476 &	0.825 &	115.844	& 0.661 \\
Sundial         & 0.778	& 655.847 &	0.491 &	1316.141 &	0.630 &	177.306	& 0.633 \\
Panda           & 0.810 & 525.939 &	0.304 &	1360.377 &	0.647 &	171.515	& 0.652 \\
TiReX           & \underline{0.739}	& 320.896 &	\bf 0.644 &	\bf 1072.251 &	0.809 &	116.183	& 0.465 \\
Chronos-2       & \bf 0.698	& 501.283 &	\underline{0.617} &	\underline{1123.497} &	0.808 &	115.609	& 0.500 \\
\hdashline
\bf NormWear-2 (Ours) & 0.842 &	\underline{167.303} & 0.608 & 1134.325 & \underline{0.877} & \underline{94.520}	& \bf 0.457 \\
\bf NormWear-2 Insight only & 0.883 &	\bf 145.560 & 0.600 & 1161.489 & \bf 0.894 & \bf 86.262 & \underline{0.461} \\

\hline
\multicolumn{8}{c}{\textbf{Minute Level (PMData, sport wristband daily sensing)}} \\
\hline
Naive           & 0.585 & 460.738 &	0.417 &	429.503 & 0.650 & 219.015 &	0.606 \\
Seasonal Naive  & 0.685	& 443.865 &	0.386 &	516.992 & 0.666 & 196.836 &	0.657 \\
Sundial         & 0.609 & 428.150 &	\bf 0.700 &	416.094 & 0.724 & 175.686 &	0.527 \\
Panda           & 0.642 & 459.234 &	0.637 &	430.185 & 0.717 & 175.398 &	0.558 \\
TiReX           & \bf 0.477 & 438.614 &	0.526 &	386.534 & \underline{0.733} & 183.006 &	0.523 \\
Chronos-2       & \underline{0.481} & 443.463 &	0.493 &	398.692 & 0.716 & 190.676 &	0.541 \\
\hdashline
\bf NormWear-2 (Ours) & 0.653 &	\bf 293.541 & \underline{0.652} & \bf 361.957 &	\bf 0.801 &	\underline{141.341} & \bf 0.466 \\
\bf NormWear-2 Insight only & 0.705 & 332.162 & 0.646 & 381.700 & \bf 0.801 & \bf 140.976 & \underline{0.494} \\

\hline
\multicolumn{8}{c}{\textbf{Minute Level (CGMacros, biofluidic sensing)}} \\
\hline
Naive           & 0.761	& 622.353 & 0.520 & 524.629	& 0.609	& 230.554 & 0.709 \\
Seasonal Naive  & 0.959	& 681.113 & 0.466 & 605.890	& 0.703	& 192.298 & 0.778 \\
Sundial         & 0.763	& 479.786 & 0.731 & 458.067	& 0.684	& 193.140 & 0.590 \\
Panda           & 0.807	& 465.601 & 0.651 & 441.006	& 0.700	& 181.970 & 0.594 \\
TiReX           & \underline{0.720}	& 491.055 & 0.665 & 423.279	& 0.722	& 181.486 & 0.571 \\
Chronos-2       & \bf 0.676	& 453.967 & 0.683 & 414.460	& 0.723	& 181.411 & 0.548 \\
\hdashline
\bf NormWear-2 (Ours)        & 0.851 & \bf 239.749 & \underline{0.740} & \bf 377.904 & \underline{0.822} & \underline{131.524} & \bf 0.474 \\
\bf NormWear-2 Insight only & 0.928 & \underline{263.617} & \bf 0.743 & \underline{383.388} & \bf 0.830 & \bf 127.924 & \underline{0.492} \\

\hline
\multicolumn{8}{c}{\textbf{Quarter Hour Level (Shanghai Diabetes, biofluidic sensing)}} \\
\hline
Naive           & 0.875	& 246.373 &	0.514 &	157.369 & 0.744 & 204.183 &	0.801 \\
Seasonal Naive  & 0.938	& \underline{143.211} & 0.673 & 130.531	& 0.867	& 126.337 & 0.611 \\
Sundial         & 0.884	& 207.405 & 0.692 & 139.755	& 0.794	& 172.074 & 0.693 \\
Panda           & 0.953	& 154.500 & 0.692 & 125.636	& 0.846	& 140.096 & 0.618 \\
TiReX           & \bf 0.817	& 180.460 & \underline{0.731} & 124.321	& 0.822	& 155.455 & 0.617 \\
Chronos-2       & \underline{0.856}	& 218.798 & \bf 0.759 & 130.564	& 0.842	& 144.237 & 0.657 \\
\hdashline
\bf NormWear-2 (Ours) & 1.008 &	\bf 129.214	& 0.725	& \bf 118.784 & \underline{0.880} & \underline{119.986}	& \bf 0.578 \\
\bf NormWear-2 Insight only & 1.090 & 144.634 & 0.726	& \underline{120.966} & \bf 0.882 & \bf 119.306 & \underline{0.608} \\

\hline
\multicolumn{8}{c}{\textbf{Hour Level (KidneyDialysis, medical device sensing)}} \\
\hline
Naive           & \underline{0.832} & 136.564 &	0.508 &	102.106 & 0.749 & 211.231 &	0.752 \\
Seasonal Naive  & 1.037	& 114.633 &	0.660 &	87.619 & \bf 0.847 & \bf 148.814 &	0.665 \\
Sundial         & 0.846	& 126.489 &	0.729 &	90.928 & 0.791 & 182.420 & 0.662 \\
Panda           & 0.867 & 104.077 &	\bf 0.743 &	80.485 & 0.814 & 168.567 & 0.600 \\
TiReX           & \bf 0.808 & 109.522 &	0.725 &	80.086 & 0.812 & 170.383 & 0.600 \\
Chronos-2       & 0.835 & 104.425 &	0.748 &	80.286 & 0.832 & 158.743 & 0.589 \\
\hdashline
\bf NormWear-2 (Ours) & 0.886 &	\underline{87.886} & \underline{0.741} & \underline{79.923} & 0.834 & 156.169 &	\underline{0.575} \\
\bf NormWear-2 Insight only & 0.921 &	\bf 87.623 & \underline{0.741} & \bf 78.217 & \underline{0.839}	& \underline{152.939} & \bf 0.574 \\

\hlineB{2}
\end{tabular}
}
\label{tab:main_results}
\vspace{-4mm}
\end{table*}

\begin{figure*}[t]
\vspace{-4mm}
\centering
\includegraphics[width=1.0\textwidth]{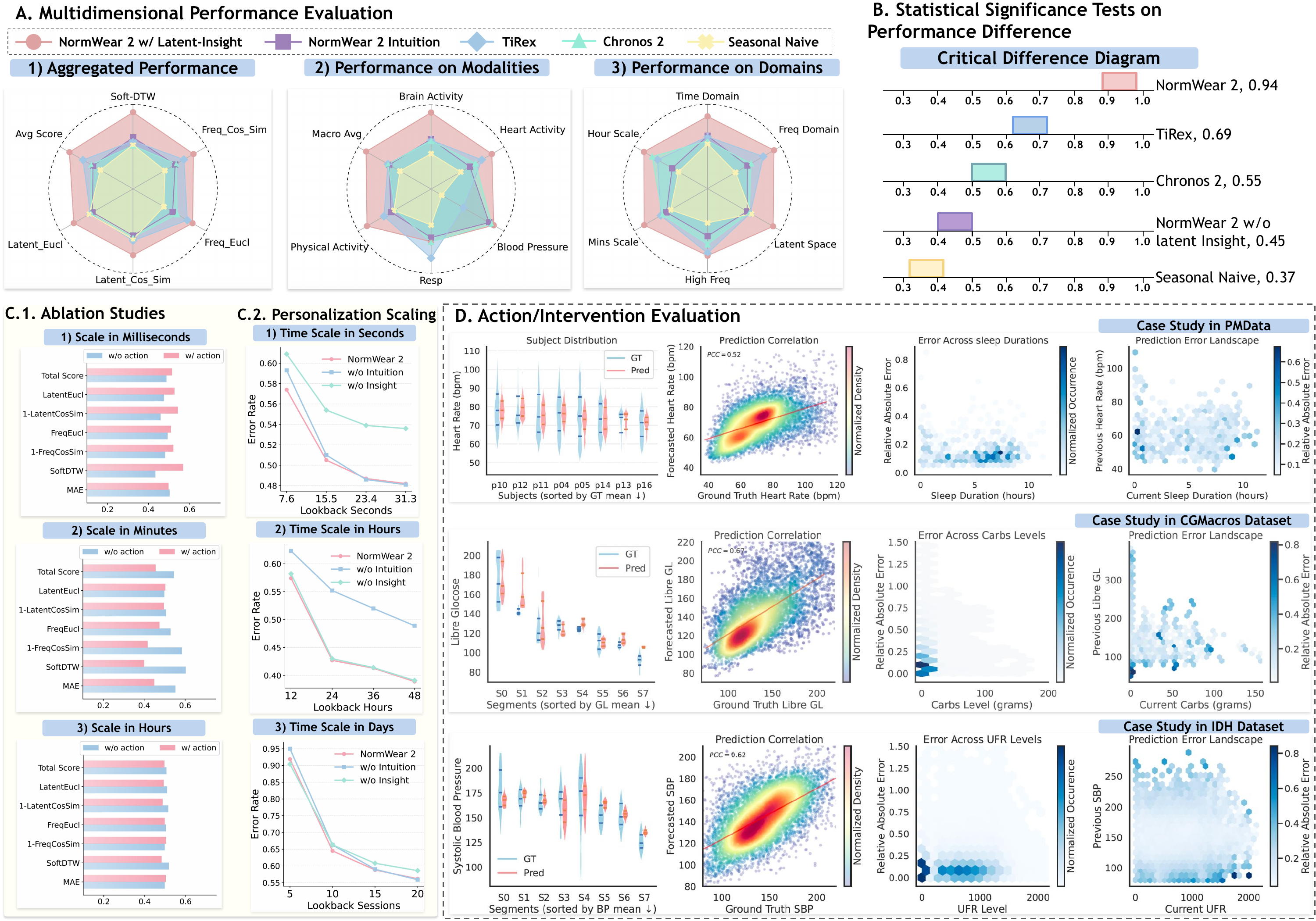}
\vspace{-7mm}
\caption{\textbf{Quantitative Results.} 
(A) Relative performance comparison of generative forecasting quality. 
(B) Statistical test result shows that the models performances are significantly different. 
(C) Overview of the ablation study results. 
(D) Model inference behavior under varied actions.  
}
\label{fig:quantitative_results}
\vspace{-4mm}
\end{figure*}

\subsection{Preliminary Exploration on Chaos-Balance-Aware Pretraining}
\vspace{-2mm}
To investigate the effect of chaos-balance-aware pretraining, we evaluate models on both generative and downstream tasks, including forecasting, simulation, classification, and regression. Generation quality is measured by MAE/MSE at this stage, while downstream performance is assessed using task-specific metrics following \cite{normwear}.  
To isolate the role of dataset balance, we construct controlled pretraining subsets with identical data size ($10^5$ samples) but different balance scores, where balance is quantified using weighted Shannon entropy and granularity-based diversity measures over clustered dynamical systems (Appendix~\ref{app:entropy_granular}). 
As shown in Figure~\ref{fig:balance_results} panel B, models pretrained on more balanced subsets consistently achieve lower generative error across evaluation settings.

We then compare a smaller but more balanced corpus ($10^5$ samples, balance score 0.73) against a larger but less balanced corpus ($2\times10^5$ samples, balance score 0.60), while keeping architecture and optimization settings fixed. As shown in the lower half of panel B in Figure \ref{fig:balance_results}, despite the reduced data size, the more balanced pretraining set yields better performance across multiple scenarios. 
%
We further observe that data scaling and balance interact synergistically: when increasing pretraining data size under fixed model capacity, more balanced datasets exhibit faster reductions in test-time generative error. Together, these preliminary findings suggest that chaos-balance-aware curation tends to be the one of the core aspects that can be intuitively identified that provides the model a decent generative forecasting quality. We provide supplementary empirical details in Appendix \ref{app:eval-balance-all}. 

Finally, we evaluate chaos awareness using the Forced Van der Pol as the paradigm system (Figure \ref{fig:balance_results}C). Its bifurcation diagram shows that in stable regions the system converges to one or two local optima, while increasing the forcing parameter drives the system into chaos; for example, when $a \geq 0.9875$, the optima become highly dispersed, indicating strong chaoticity. We compare forecasting performance between a chaos-balanced pre-trained model and a sensing-only model by conditioning on the first 2048 of 4096 sampled steps and predicting the remaining steps. The chaos-balanced model performs well in stable regions and degrades mainly after the onset of chaos, while still maintaining reasonable accuracy overall. In contrast, the sensing-only model performs poorly across all regimes (its error is scaled by 50\% in the figure for visibility). T-SNE visualization further shows that the chaos-balanced model clearly separates systems with high and low Lyapunov exponents in latent space. These results suggest that chaos-balanced pre-training improves the model’s understanding of dynamical structure and chaotic transitions, leading to better forecasting performance.

\vspace{-2mm}
\subsection{Quantitative Results of Forecasting on Physiological Signals}
\vspace{-2mm}

Table~\ref{tab:main_results} summarize the main forecasting results on physiological signals under our multidimensional evaluation framework. To contextualize performance, we compare against several baselines, including running-average prediction (Naive), repetition of the dominant periodic pattern (Seasonal Naive), state-of-the-art time-series foundation models \citep{sundial,tirex,chronos2}, and the state-of-the-art dynamical-system foundation model \citep{panda}. 
The digital healthcare downstream inference results shown in Table~\ref{tab:main-res-downstream} primarily serve as an assessment of representation quality, providing additional context on whether models that achieve stronger generative forecasting performance also preserve the quality of their learned representations for downstream tasks. 
Because forecasting quality is evaluated from multiple complementary perspectives, no single method consistently achieves the best score on every individual metric. Instead, different methods exhibit distinct strengths under different criteria, underscoring the importance of multi-faceted evaluation. NormWear-2 achieves the strongest overall performance by offering the most favorable trade-off across metrics. 
To assess the robustness of this advantage, we conduct Conover post hoc tests on model rankings across temporal resolutions, sensing modalities, evaluation metrics, and datasets. As shown in Figure~\ref{fig:quantitative_results}A-B, 
NormWear-2 maintains the best overall ranking with statistical significance over prior state-of-the-art baselines.

\vspace{-2mm}
\subsection{Counterfactual Validation on Model's Action Awareness}
\vspace{-2mm}
To qualitatively assess whether the learned transition model captures action-conditioned physiological dynamics, we conduct analysis on the healthcare datasets. In the kidney dialysis scenario for example, the action variables correspond to dialysis machine control parameters prescribed to the patient and adjusted by clinicians during treatment when intolerance symptoms arise (e.g., headache, dyspnea, or chest discomfort). Among these parameters, ultrafiltration rate (UFR) is the primary intervention variable in practice. In severe cases, UFR may be reduced substantially or completely shut down to pause fluid removal. Clinically, lower UFR generally reduces the risk of adverse symptoms, but also prolongs the dialysis session, creating an inherent treatment trade-off.
While optimizing dialysis control policy is beyond the scope of this work and would require a dedicated reinforcement learning framework, our objective here is to model physiological state transitions conditioned on observed clinical actions. In particular, we focus on forecasting future systolic blood pressure (SBP), one of the most critical biomarkers monitored during dialysis.

Figure~\ref{fig:quantitative_results}D provides qualitative evidence that the model learns clinically meaningful action-conditioned dynamics. First, the predicted SBP distribution remains well aligned with the ground-truth physiological range, indicating stable transition modeling. Second, predicted SBP exhibits a moderate positive linear relationship with the observed next-step SBP across representative cases, suggesting that the model preserves subject-specific physiological trends. Most importantly, when evaluated across varying UFR settings, the model maintains low relative prediction error in the majority of cases (typically below 0.2), indicating robust transition estimation under different treatment configurations.

To further inspect action sensitivity, we visualize the prediction error landscape across varying UFR levels and prior SBP conditions. Across most evaluated cases, prediction error remains consistently low and is dominated by regions below 0.1, suggesting that the learned dynamics remain stable under diverse action-state combinations. Notably, several subjects exhibit localized regions of elevated error, reflecting more complex or atypical physiological responses that are harder to capture with population-level dynamics alone.
These observations suggest that NormWear-2 captures meaningful action-dependent physiological trends and exhibits sensitivity to intervention changes across a broad range of treatment settings. At the same time, the observed high-error outlier cases highlight the importance of personalization: accurately modeling such subjects likely requires adapting the transition dynamics as more subject-specific historical data become available. 
Similar analysis results in PMData and CGMacros, representing daily-life and point-of-care scenarios, are also presented in Figure \ref{fig:quantitative_results} panel D. More analysis examples are shown in Appendix \ref{app:action_aware_studies_more}.

\begin{table*}[t]
\centering
\small
\caption{Evaluate compatibility with representative alternative backbone models. Scores are reported as (Score w/o latent insight, Score w/ latent insight). Improvements are highlighted in bold when significantly better than the non–latent-insight method.
``Attn" stands for Attention.
}
\renewcommand{\arraystretch}{1.2}
\resizebox{0.99\textwidth}{!}{
\begin{tabular}{l|c|c|c|c|c|c|c}
\hlineB{2}
Backbone Alternatives 
& MAE $\downarrow$ 
& Soft-DTW $\downarrow$ 
& FreqCosSim $\uparrow$ 
& FreqEucl $\downarrow$ 
& LatentCosSim $\uparrow$ 
& LatentEucl $\downarrow$ 
& \bf Final Score $\downarrow$ \\
\hline

\multicolumn{8}{c}{\textbf{Millisecond Level (VitalDB, wearable and medical device sensing)}} \\
\hline

NormWear-2 Univariate
& 0.888 \; \bf 0.843
& 301.595 \; \bf 233.677
& 0.475 \; \bf 0.525
& 1412.252 \; \bf 1251.776
& 0.812 \; 0.767
& 116.814 \; 134.193
& 0.569 \; \bf 0.528 \\

NormWear-2 [CLS]-Attn
& 0.879	\; \bf 0.851
& 607.640 \; \bf 198.523
& 0.452 \; \bf 0.611
& 1387.582	\; \bf 1138.682
& 0.716	\; \bf 0.860
& 151.818 \; \bf 101.977
& 0.640 \; \bf 0.468 \\

NormWear-2 LeWM-JEPA
& 0.813	\; 0.819
& 298.418 \; \bf 283.110
& 0.525 \; \bf 0.535
& 1220.297 \; \bf 1206.771
& 0.753 \; 0.743
& 139.283 \; 142.702
& 0.531 \; \bf 0.528 \\


NormWear-2 SFT
& 0.733	\; 0.813
& 278.402 \; \bf 191.238
& 0.616 \; 0.616
& 1085.561 \; 1130.278
& 0.759 \; \bf 0.858
& 131.326 \; \bf 102.616
& 0.475 \; \bf 0.459 \\


\hline
\multicolumn{8}{c}{\textbf{Minute Level (PMData, sport wristband daily sensing)}} \\
\hline

NormWear-2 Univariate
& 0.625	\; 0.633
& 390.088 \; \bf 284.748
& 0.581 \; \bf 0.619
& 390.470 \; \bf 376.389
& 0.734 \; \bf 0.746
& 172.824 \; \bf 170.531
& 0.527 \; \bf 0.489 \\

NormWear-2 [CLS]-Attn
& 0.939	\; \bf 0.714
& 434.040 \; \bf 260.89
& 0.614 \; \bf 0.711
& 439.472 \; \bf 366.758
& 0.713 \; \bf 0.777
& 176.767 \; \bf 154.105
& 0.611 \; \bf 0.468 \\

NormWear-2 LeWM-JEPA
& 0.614	\; 0.661
& 297.461 \; \bf 279.448
& 0.688 \; \bf 0.691
& 373.915 \; 380.756
& 0.757	\; 0.743
& 156.661 \; 160.783
& 0.471 \; 0.480\\



\hline
\multicolumn{8}{c}{\textbf{Minute Level (CGMacros, biofluidic sensing)}} \\
\hline

NormWear-2 Univariate
& 1.029	\; \bf 0.967
& 322.494 \; \bf 245.383
& 0.691 \; \bf 0.710
& 427.976 \; \bf 400.790
& 0.764	\; 0.756
& 150.677 \; 157.238
& 0.568 \; \bf 0.527 \\

NormWear-2 [CLS]-Attn
& 1.069	\; \bf 0.895
& 438.332 \; \bf 279.720
& 0.681 \; \bf 0.753
& 502.969	\; \bf 388.924
& 0.707	\; \bf 0.774
& 177.405 \; \bf 150.909
& 0.653 \; \bf 0.506 \\

NormWear-2 LeWM-JEPA
& 0.770	\; 0.798
& 248.940 \; \bf 237.531
& 0.747 \; 0.712
& 402.047	\; 412.743
& 0.763	\; 0.753
& 152.345 \; 156.4
& 0.487 \; 0.502 \\



\hline
\multicolumn{8}{c}{\textbf{Quarter Hour Level (Shanghai Diabetes, biofluidic sensing)}} \\
\hline

NormWear-2 Univariate
& 0.946	\; 1.002
& 173.465 \; \bf 130.461
& 0.652 \; \bf 0.724
& 128.306	\; \bf 117.504
& 0.815 \; \bf 0.883
& 157.896 \; \bf 119.879
& 0.654 \; \bf 0.576 \\

NormWear-2 [CLS]-Attn
& 0.861	\; 0.954
& 189.533 \; \bf 142.906
& 0.733 \; 0.697
& 133.683	\; \bf 122.302
& 0.802	\; \bf 0.852
& 166.080 \; \bf 138.035
& 0.654 \; \bf 0.601 \\

NormWear-2 LeWM-JEPA
& 1.002	\; 1.030
& 186.225 \; \bf 135.697
& 0.685 \; \bf 0.701
& 125.241	\; \bf 118.791 
& 0.818	\; \bf 0.860
& 156.359 \; \bf 132.92 
& 0.663 \; \bf 0.598 \\



\hline
\multicolumn{8}{c}{\textbf{Hour Level (KidneyDialysis, medical device sensing)}} \\
\hline

NormWear-2 Univariate
& 0.874	\; 0.920
& 109.323 \; \bf 81.342
& 0.690 \; \bf 0.741
& 82.866 \; \bf 76.013
& 0.789 \; \bf 0.847
& 182.831 \; \bf 151.855
& 0.630 \; \bf 0.559 \\

NormWear-2 [CLS]-Attn
& 0.864	\; 0.916
& 111.051 \; \bf 85.729
& 0.723 \; \bf 0.738
& 83.154 \; \bf 77.116
& 0.802 \; \bf 0.842
& 180.434 \; \bf 154.665
& 0.623 \; \bf 0.569 \\

NormWear-2 LeWM-JEPA
& 0.900	\; 0.909
& 93.222 \; \bf 84.988
& 0.753 \; 0.749
& 78.609 \; 78.020
& 0.825 \; \bf 0.833
& 159.991	\; \bf 155.750
& 0.581 \; \bf 0.569 \\


NormWear-2 SFT
& 0.789	\; 0.889
& 97.428 \; \bf 81.959
& 0.764 \; \bf 0.730
& 86.400 \; \bf 78.644
& 0.818 \; \bf 0.843
& 167.1 \; \bf 152.037
& 0.589 \; \bf 0.564 \\

\hlineB{2}
\end{tabular}
}
\label{tab:ablations}
\vspace{-4mm}
\end{table*}

\vspace{-2mm}
\subsection{Ablation Studies: Personalized Scaling and Compatibility with Alternative Backbones}
\vspace{-2mm}

We conduct an ablation study to evaluate the degree of personalization enabled by NormWear-2 through its latent dynamical transition modeling mechanism. As shown in Figure~\ref{fig:quantitative_results}C.2, increasing the amount of available historical physiological records consistently improves forecasting performance across multiple temporal resolutions. This trend suggests that NormWear-2 can effectively leverage accumulating user-specific history to achieve progressively better personalized forecasting, highlighting its potential for continual improvement in real-time monitoring deployments. 
%
Moreover, as summarized in Table~\ref{tab:ablations}, our proposed latent dynamical transition modeling approach consistently improves performance when integrated with a diverse set of representative backbones, including the univariate backbone \citep{patchtst}, the CLS-attention backbone \citep{normwear}, the LE-world model JEPA \citep{maes2026leworldmodel}, and supervised fine-tune (SFT) backbone in scenarios with large population. Despite their architectural differences and varying inductive biases, equipping each backbone with our transition modeling mechanism leads to significant and consistent gains across multiple evaluation metrics. This demonstrates that the effectiveness of our approach is largely backbone-agnostic, highlighting its strong compatibility and generalizability in enhancing temporal representation learning across varied modeling paradigms. 



\vspace{-1mm}
\section{Discussion and Conclusion} \label{sec:conclude}
\vspace{-2mm}

\textbf{Limitations and Future Work.} While our framework incorporates action-like mechanisms through state transition adaptation in latent space, it does not yet formulate these interactions within a general reinforcement learning paradigm with explicit reward optimization. In many real-world healthcare scenarios, decision-making involves optimizing long-term outcomes under uncertainty, where actions should be guided by well-defined objectives rather than implicit adaptation alone. Extending our latent world model with a generic RL framework, where policies operate over latent states and are trained to optimize clinically meaningful rewards, represents a key direction for future work.

\textbf{Broader Impact.} This work contributes to the development of general-purpose, predictive modeling frameworks for physiological signals, with potential applications in digital health, continuous monitoring, and personalized medicine. By moving beyond task-specific models toward latent representations that capture underlying dynamics, our approach may enable earlier detection of health deterioration, improved forecasting of disease progression, and more adaptive health management systems. More broadly, this work highlights the value of combining dynamical systems theory with modern representation learning as a foundation for next-generation intelligent healthcare systems. 



\section*{Ethics Statement}
This study contains applications in the field of healthcare. We ensured that all the data being used during pretraining and evaluations were either made publicly available by the original authors, or acquired from studies with IRB approval, and all these works were cited properly. Details of the datasets are specified in Appendix \ref{app:datasets}. 

\section*{Reproducibility Statement}
The pretraining code follows prior work by \citet{normwear}. The code of model, pretrain data, and latest checkpoints are publicly available on Hugging Face under the \textbf{\textsc{NormWear} Collection:} \href{https://huggingface.co/collections/mosaic-laboratory/normwear}{\texttt{mosaic-laboratory/normwear}}
. The sources of data are described and properly referenced through out the paper. 

\bibliography{references}
\bibliographystyle{nips}

\clearpage
\appendix


\section{Datasets} \label{app:datasets}

\begin{table*}[htbp!]
\vspace{-2mm}
\centering
\small
\caption{Lengths of input context and forecast prediction for evaluations.}
\renewcommand{\arraystretch}{1.0}
\resizebox{\textwidth}{!}{
\begin{tabular}{
    l
    c
    c
    c
    c
}
\hlineB{2}
\textbf{Dataset} & 
\textbf{Max Observed Context} & 
\textbf{Min Observed Context} & 
\textbf{Prediction Length} &
\textbf{\# Test Samples} \\
\hline

\makecell[l]{VitalDB \\ \citep{vitaldb}}
& 2,032 (31.3 seconds)
& 496 (7.6 seconds)
& 2,064 (31.8 seconds)
& 62,910
\\
\hline
\makecell[l]{PMData \\ \citep{pmdata}} 
& 2,880 (2 days)
& 720 (half a day)
& 720 (half a day)
& 886
\\
\hline
\makecell[l]{CGMacros \\ \citep{cgmacros}} 
& 2,880 (2 days)
& 720 (half a day)
& 720 (half a day)
& 173
\\
\midrule 
\makecell[l]{Shanghai Diabetes \\ \citep{shanghai_diabete}} 
& 192 (2 days)
& 96 (1 day)
& 192 (2 days)
& 223
\\
\hline
\makecell[l]{KidneyDialysis \\ \citep{idh}} 
& \makecell[c]{186 measurements \\ (20 dialysis session)} 
& \makecell[c]{46 measurements \\ (5 dialysis session)}
& \makecell[c]{37 measurements \\ (4 dialysis session)}
& 22,811
\\

\hlineB{2}
\end{tabular}
}
\label{tab:dataset_lengths}
\vspace{-3mm}
\end{table*}

\begin{table}[htbp!]
\centering
\caption{
\textbf{Characteristic of the Analyzed Population in the kidney dialysis (KidneyDialysis) dataset.}
SD, standard deviation; 
BMI, body mass index; 
min, minute(s).
}
\label{tab:idh_dataset_stats}
\begin{tabular}{p{9cm} r}
\toprule
\multicolumn{2}{l}{\textbf{Basic Statistics}} \\
\midrule
Total number of patients & 1,452 \\
Age (years), mean (SD) &  65 (15) \\
BMI (Kg/m$^2$), mean (SD) & 28.8 (8.4) \\
Dialysis vintage, years, mean (SD) & 2.3 (1.6) \\
Gender, n ($\%$) & \\
\;\;Male & 846 (58) \\
\;\;Female & 606 (42) \\
Race, n ($\%$) & \\
\;\;White & 789 (54) \\
\;\;African American & 247 (17) \\
\;\;Hispanic & 174 (12) \\
Number measurements per session, mean (SD) & 9.3 (1.2) \\
Measurement gap, mean (SD), minutes & 26.3 (14.7) \\
Measurement gap distribution (5\%, 25\%, 50\%, 75\%, 95\%), min & (3, 17, 30, 31, 53) \\
Total number of sessions & 211,397 \\
Total number of samples & 1,965,389 \\
\midrule
\multicolumn{2}{l}{\textbf{Validation Statistics}} \\
\midrule
Number of subjects (for\_SFT\_study - testing) & 1,161 \,-\, 291 \\
\bottomrule
\end{tabular}
\end{table}

\textbf{VitalDB} is an open-access dataset designed to support machine learning research in anesthesia and perioperative monitoring. It contains high-resolution waveform and numeric biosignal data collected from 6,388 surgical cases, covering 196 intraoperative monitoring parameters, 73 perioperative clinical variables, and 34 laboratory time-series parameters. The dataset was created to address the shortage of large-scale biosignal datasets for developing predictive and analytical models in anesthesiology and patient monitoring. The detailed data statistics were presented in the original paper \citep{vitaldb}. During the evaluation, data from $20\%$ (1,278/6,388) of total surgery cases are left out for methods assessment, and the rest of the cases are reserved for supervised fine-tune (SFT) study.

\textbf{PMData} is a multimodal dataset that combines daily lifelogging information with sports and physical activity records to support health and performance analysis. Data were collected from 16 participants over five months using Fitbit Versa 2 smartwatches, the PMSys smartphone application, and Google Forms. It enables research on relationships between exercise, sleep, body weight, and athletic performance, allowing both lifestyle prediction and sports-oriented analytics. The detailed data statistics were presented in the original paper \citep{pmdata}.

\textbf{CGMacros} is a multimodal dataset developed for personalized nutrition and glucose response analysis. It includes continuous glucose monitor data, food macronutrient records, meal photographs, physical activity, demographic information, blood biomarkers, and gut microbiome profiles from 45 participants with different metabolic conditions. Participants were monitored for ten consecutive days in free-living settings while consuming meals with controlled macronutrient compositions. The detailed data statistics were presented in the original paper \citep{cgmacros}.

\textbf{Shanghai Diabetes} dataset consists of two publicly available datasets created to support data-driven diabetes management research for both Type 1 and Type 2 diabetes patients. It includes data from 12 Type 1 diabetes patients and 100 Type 2 diabetes patients collected under real-life conditions in Shanghai, China. The dataset provides clinical characteristics, laboratory measurements, medications, continuous glucose monitoring records, and daily dietary information for developing glucose prediction and disease management models. The detailed data statistics were presented in the original paper \citep{shanghai_diabete}.

\textbf{KidneyDialysis \citep{idh}} 
is a retrospective real-world dataset, with valid IRB approval, developed for studying physiological and treatment-related patterns during maintenance hemodialysis. The dataset is acquired through a signed consent from the original authors. It includes 1,452 patients who received in-center hemodialysis across 17 Sanderling Care clinics in the United States between 2013 and 2024, comprising a total of 211,397 dialysis treatment sessions. High-resolution physiological and dialysis machine data were automatically transmitted from dialysis machines to the secure PEARL data management platform, including heart rate, body temperature, blood pressure, blood flow rate, dialysate parameters, ultrafiltration rate, cumulative fluid removal, and treatment duration. Because the original full data statistics were not reported in the original poster, we present the detailed data statistics in Table \ref{tab:idh_dataset_stats} based on our own analysis process. 
During the evaluation, data from $20\%$ (291/1,452) of total patients are left out for methods assessment, and the rest of the cases are reserved are reserved for supervised fine-tune (SFT) study.

\subsection{Pretrain Data}
We leverage the datasets released by \citet{normwear} for 18 wearable downstream tasks, \citet{panda}'s evaluation set for evaluating the generative quality on real-world chaotic system, \citet{batterylife}'s datasets for the evaluation on battery test time series, and \citet{autoformer}'s datasets for evaluation on civil monitoring forecasting tasks. The pre-train datasets are completely disjoint collections of multivariate time series data, with statistics presented in Table \ref{tab:pretrain-data}. 

\begin{table}[ht]
    \centering
    \renewcommand{\arraystretch}{1.0} 
    \setlength{\tabcolsep}{4pt} 
        \centering
        \caption{\textbf{Pretrain Datasets.}}
        \scalebox{0.8}{
        \begin{tabular}{lccc}
        \hlineB{2}
        \hlineB{2}
        \bf Datasets & \bf Sequence Length & \bf \# Samples & \bf \# Variates 
        \\
        \hline
        \citet{normwear} & 390 & $2.3 \times 10^5$ & \{2,3,4,6\} 
        \\
        \citet{panda} & 4096 & $1.0 \times 10^5$ & \{3,4,6\} 
        \\
        Aggregated Benchmark & \{390, 4096\} & $2.2 \times 10^5$ & \{2,3,4,6\} 
        \\
        Balanced Benchmark & \{390, 4096\} & $1.0 \times 10^5$ & \{2,3,4,6\} 
        \\
        \hlineB{2}
        \hlineB{2}
        \end{tabular}
        }
        \label{tab:pretrain-data}
\end{table}

\subsection{\edit{Evaluation Datasets for Preliminary Chaos-Aware Pre-train Experiments}} \label{app:test-info}
\edit{
The detailed statistics of the downstream tasks including the battery state of health (SOH) prediction and digital health downstream tasks were presented in prior works in the literature: \cite{batterylife} and \cite{normwear} respectively. The data statistics of the genrative tasks are summarized in Table \ref{tab:dataset-stats} as shown below. 
}

\begin{table*}[htpb!]
\centering
\caption{\edit{Statistics of generative testing datasets used in our experiments.}}
\resizebox{\textwidth}{!}{
\begin{tabular}{lccc}
\toprule
\textbf{Dataset} & \textbf{Data Points per Channel} & \textbf{Channels} & \textbf{Domain} \\
\midrule
Panda Test Set \citep{panda} & $3.4\times 10^7$ & $[3, 6]$ & Chaotic System \\
WESAD  \citep{normwear}  & $4.2\times 10^6$ & $[6]$    & Wearable Sensing \\
HUST   \citep{batterylife}  & $1.0\times 10^6$ & $[3]$    & Battery Health and Material Test \\
CALB   \citep{batterylife}    & $1.6\times 10^4$ & $[3]$    & Battery Health and Material Test \\
Na-ion  \citep{batterylife}    & $2.0\times 10^4$ & $[3]$    & Battery Health and Material Test \\
Zn-coin  \citep{batterylife}   & $6.4\times 10^5$ & $[3]$    & Battery Health and Material Test \\
ETT   \citep{sundial}   & $1.7\times 10^5$ & $[7]$    & Civil Monitoring \\
Weather \citep{sundial} & $5.2\times 10^4$ & $[21]$   & Civil Monitoring \\
illness  \citep{sundial}  & $1.0\times 10^3$ & $[7]$    & Civil Monitoring \\
Exchange Rate  \citep{sundial}  & $7.2\times 10^3$ & $[8]$    & Civil Monitoring \\
\bottomrule
\end{tabular}
}
\label{tab:dataset-stats}
\end{table*}

\clearpage
\section{Inspecting Downstream Representation Quality}
\begin{table*}[htbp!]
\vspace{-4mm}
\scriptsize
\centering
\caption{
Inspect the quality of learned embedding representations on digital health applications under linear probing evaluation. The signal name, in column ``Modality-Specific", following each performance score denotes the model specialized for that modality: PPG \citep{papagei}, ECG‑FM \citep{ecgfm}, and EEG \citep{cbramod}.}
\renewcommand{\arraystretch}{1.2}
\resizebox{0.991\textwidth}{!}{
        \begin{tabular}{l:ccccccc}
        \hlineB{2}
        \bf \begin{tabular}{@{\hspace{-0.5\tabcolsep}}l@{\hspace{-0.5\tabcolsep}}}
            Downstream
            Tasks \\ 
        \end{tabular} &
        \bf Panda
       &
        \bf Sundial
       &
        \bf Chronos 2 
       &
        \bf TiReX
       &
        \bf Modality-Specific 
       &
        \bf \begin{tabular}{@{\hspace{-0.5\tabcolsep}}l@{\hspace{-0.5\tabcolsep}}}
            NormWear \citep{normwear}
        \end{tabular}  
        &
        \bf \begin{tabular}{@{\hspace{-0.5\tabcolsep}}l@{\hspace{-0.5\tabcolsep}}}
            Normwear 2 (Ours)
        \end{tabular}  
        \\\hline
        WESAD&
       73.187 & 70.529 & 74.414 & 73.802 & 56.656(PPG)) & \bf 76.060 & 72.524
        \\
        UCI-HAR&
        97.896 & 96.522 & 98.476 & 97.887 &-& \bf 98.954 & 98.141
        \\
        DriverFatigue&
        68.116 & 70.034 & 73.551 & 69.447 & 80.430(EEG)&\bf 74.292 & 73.178
        \\
        \hline
        \bf Activity Recognition Avg.&
        79.733&79.028& 82.147 & 80.379 & - & \bf 83.102 & 81.281
        \\
        \hline
        Epilepsy (eye open state)&
        89.958&	\bf 95.797& 94.968 & 94.351 &90.436(EEG)& 92.743 & 93.676 
        \\
        Epilepsy (eye relaxation)&
        94.085&	97.390& \bf 97.723 & 97.420 & 95.552(EEG)&94.828 & 96.639 
        \\
        Epilepsy (health area)&
       89.047&	\bf 91.812& 91.487 & 90.634 &88.065(EEG)& 88.541 & 90.079 
        \\
        Epilepsy (tumor area)&
       86.415&	\bf 91.103& 90.104 & 90.003 & 87.258(EEG) &87.197 & 88.257 
        \\
        Epilepsy (seizure)&
        98.636&	\bf 99.723& 99.541 & 99.274 &94.616(EEG)&97.053 & 99.339 
        \\
        GAMEEMO&
        55.263&	\bf 56.814& 56.795 & 55.194 &55.420(EEG)&54.937 & 54.946 
        \\
        \hline
        \bf EEG Main Tasks Avg.&
        85.567&\bf 88.773& 88.436 & 87.813 &85.225& 85.883 & 87.156
        \\
        \hline
        ECG-Abnormal&
        99.542&	99.730& 99.426 & \bf 99.737 &89.898(ECG)& 99.140 & 98.605 
        \\
         PPG-BP (HTN)&
         56.082&	55.47& 55.963 & 61.455 &61.839(PPG)&\bf 62.341 & 62.268 
        \\
        PPG-BP (DM)&
        54.992&	58.033& 59.234 & 59.647 &55.668(PPG)&55.893 & \bf 62.087 
        \\
        PPG-BP (CVA)&
        51.875&	59.514& 45.208 & 43.125 & \bf 73.125(PPG)&70.625 & 70.347 
        \\
        PPG-BP (CVD)&
        63.121&	59.275& \bf 69.223 & 54.685 &49.066(PPG)&51.773 & 62.021 
        \\
         PhysioNet EMG&
         99.948&	99.999 & 99.999 & 99.546 &-&99.216 & 99.999 
        \\
        \hline
        \bf Risk Evaluation Avg.&
        70.927&72.004& 71.509 & 69.699 & - & 73.165  & \bf 75.888
        \\
        \hline
        Noninvasive-BP&
        92.907&	90.857& 91.995 & 92.346 &90.596(PPG)& 92.420 & \bf 93.100
        \\
        PPG-Hgb&
        94.451&	94.419& 92.844 & 94.862 &\bf 94.912(PPG)&94.632 & 93.779 
        \\
        Fetal-fPCG&
       98.884&	99.082& \bf 99.105 & 98.937 & - &99.072 & 98.990 
        \\\hline
        \bf Vital Signs Avg.&
        \bf 95.414&94.786& 94.648 & 95.382 & - & 95.375 & 95.290
        \\\hline
        \bf Micro Avg.&
        81.356&82.561& 82.781 & 81.797 & - & 82.762 & \bf 83.776
        \\
        \bf Macro Avg.&
        82.910&83.648& 84.185 & 83.318 & - & 84.381 & \bf 84.904
        \\
        \hlineB{2}
        \end{tabular}
}
\label{tab:main-res-downstream}
\vspace{-4mm}
\end{table*}

\section{Visualizing Dynamical Reasoning in Forecasting}

\begin{figure*}[ht!]
\centering
\includegraphics[width=0.99\textwidth]{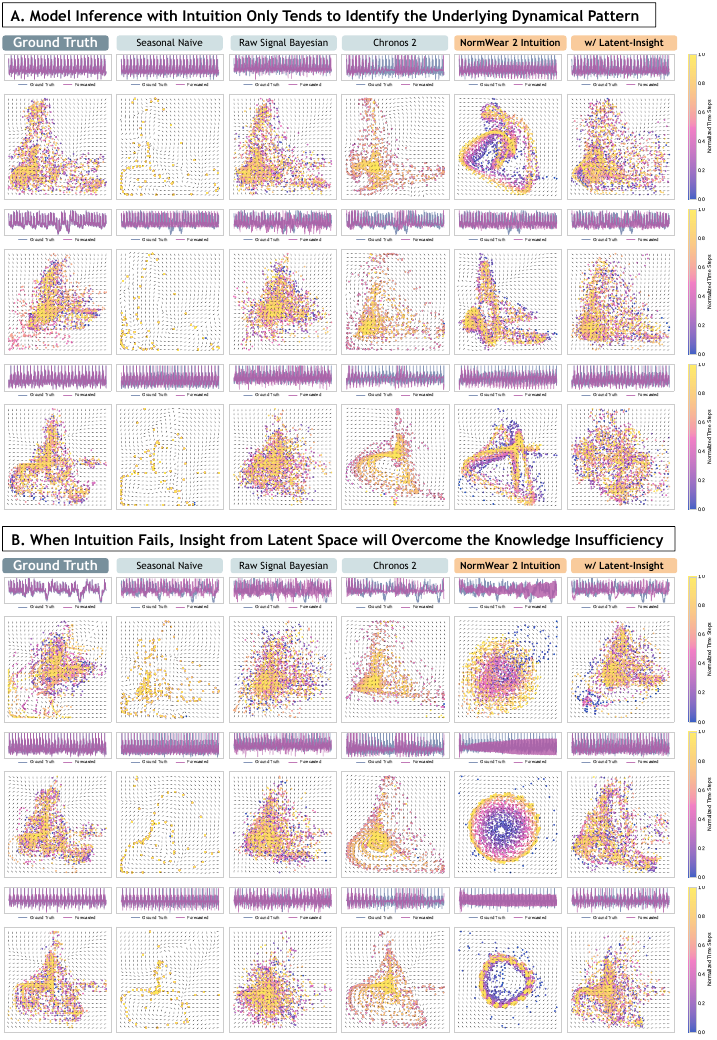}
\vspace{-3mm}
\caption{\textbf{Qualitative visualization of dynamical reasoning.}
\textbf{(A)} Successful intuition-only forecasting reconstructs phase-space trajectories with better topology.
\textbf{(B)} Failure cases where latent insight resolves ambiguity and improves forecasts.
Color indicates normalized temporal progression.}
\label{fig:latent_qualitative_results}
\end{figure*}

\begin{figure*}[ht!]
\centering
\includegraphics[width=0.99\textwidth]{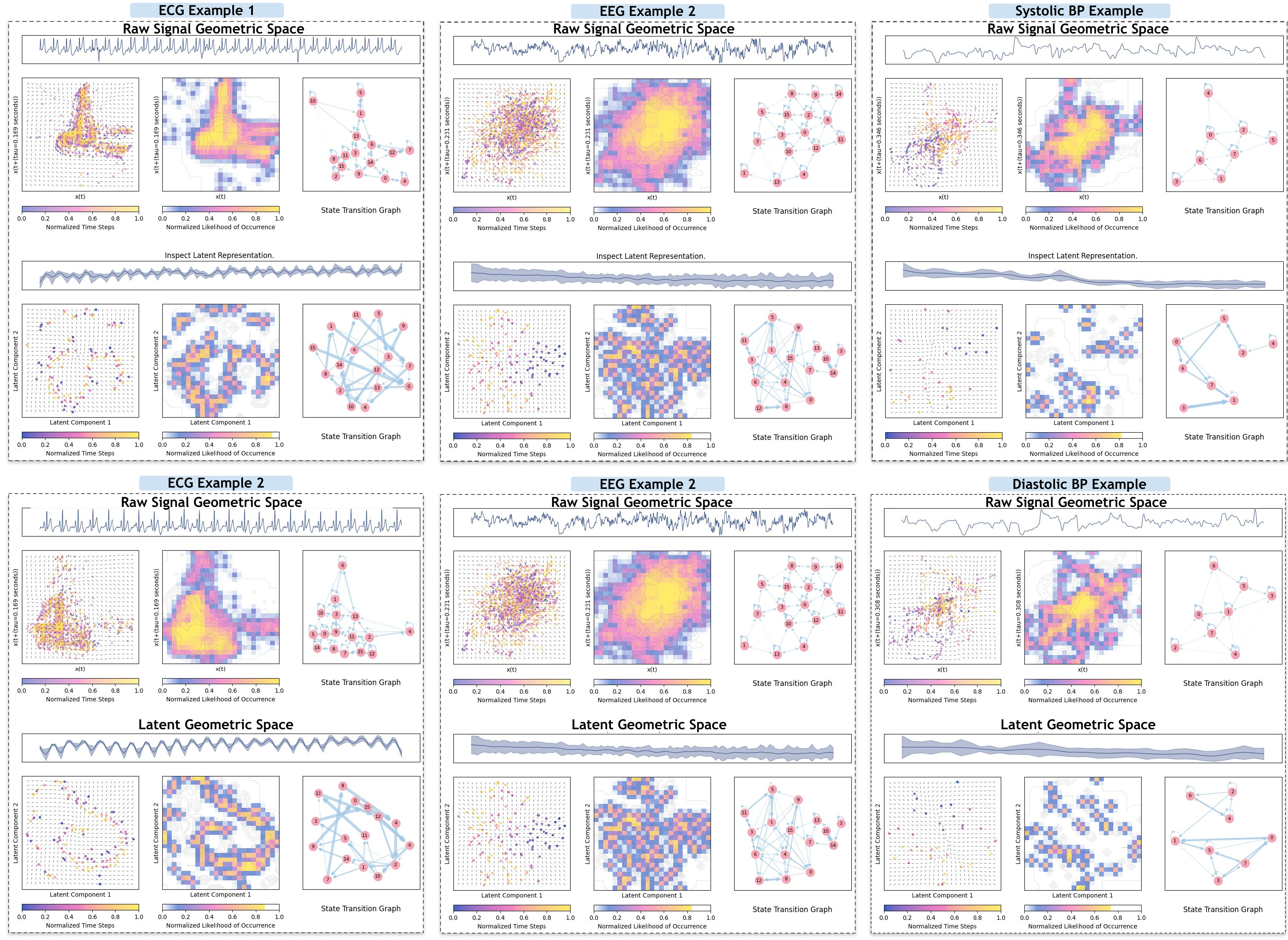}
\vspace{-3mm}
\caption{
Comparison of raw-signal dynamics and pretrained latent dynamics across representative physiological signals, showing that latent representations preserve key temporal structures including attractor geometry, occupancy distribution, and transition connectivity.
}
\label{fig:latent_qualitative_geometry}
\end{figure*}

To better understand the behaviors learned by NormWear 2 beyond quantitative metrics, we visualize both the pretrained latent representations and the forecasting dynamics in Figures \ref{fig:latent_qualitative_results} and \ref{fig:latent_qualitative_geometry}. These visualizations provide qualitative evidence that the model captures meaningful temporal structure rather than relying solely on morphological based waveform matching.
Figure \ref{fig:latent_qualitative_geometry} compares the temporal dynamics of representative physiological signals in raw signal space and pretrained latent space. Across ECG, EEG, and blood pressure examples, the latent representations preserve salient dynamical structures observed in the raw signals, including attractor geometry, temporal occupancy patterns, and state transition connectivity. This suggests that the pretrained latent space retains the underlying temporal organization of physiological dynamics while compressing the raw observations into a more structured representation.
Figure \ref{fig:latent_qualitative_results}A illustrates cases where NormWear 2 successfully infers the underlying dynamical regime from partial observations using intuition alone. Although competing baselines often produce forecasts that match local waveform statistics, NormWear 2 generates forecasts whose reconstructed phase-space trajectories demonstrate more topological details, indicating that the model tends to identifying the underlying dynamics that could inform future trajectory progression as much as possible.
Figure~\ref{fig:latent_qualitative_results}B highlights representative failure cases of intuition-only inference, where NormWear 2 did recognize a clear topological structure, but is insufficient to identify the future dynamics. In such ambiguous scenarios, incorporating latent-space insight resolves the uncertainty and produces forecasts that better recover the correct temporal evolution. This suggests that latent insight serves as a complementary mechanism to intuition-based forecasting, particularly when direct dynamical inference from partial observations is underdetermined.
These visualizations indicate that NormWear 2 learns temporally structured latent representations and leverages them to perform forecasting through dynamical pattern reasoning.

\clearpage
\section{Action Awareness Studies} \label{app:action_aware_studies_more}

\begin{figure*}[ht!]
\centering
\includegraphics[width=0.99\textwidth]{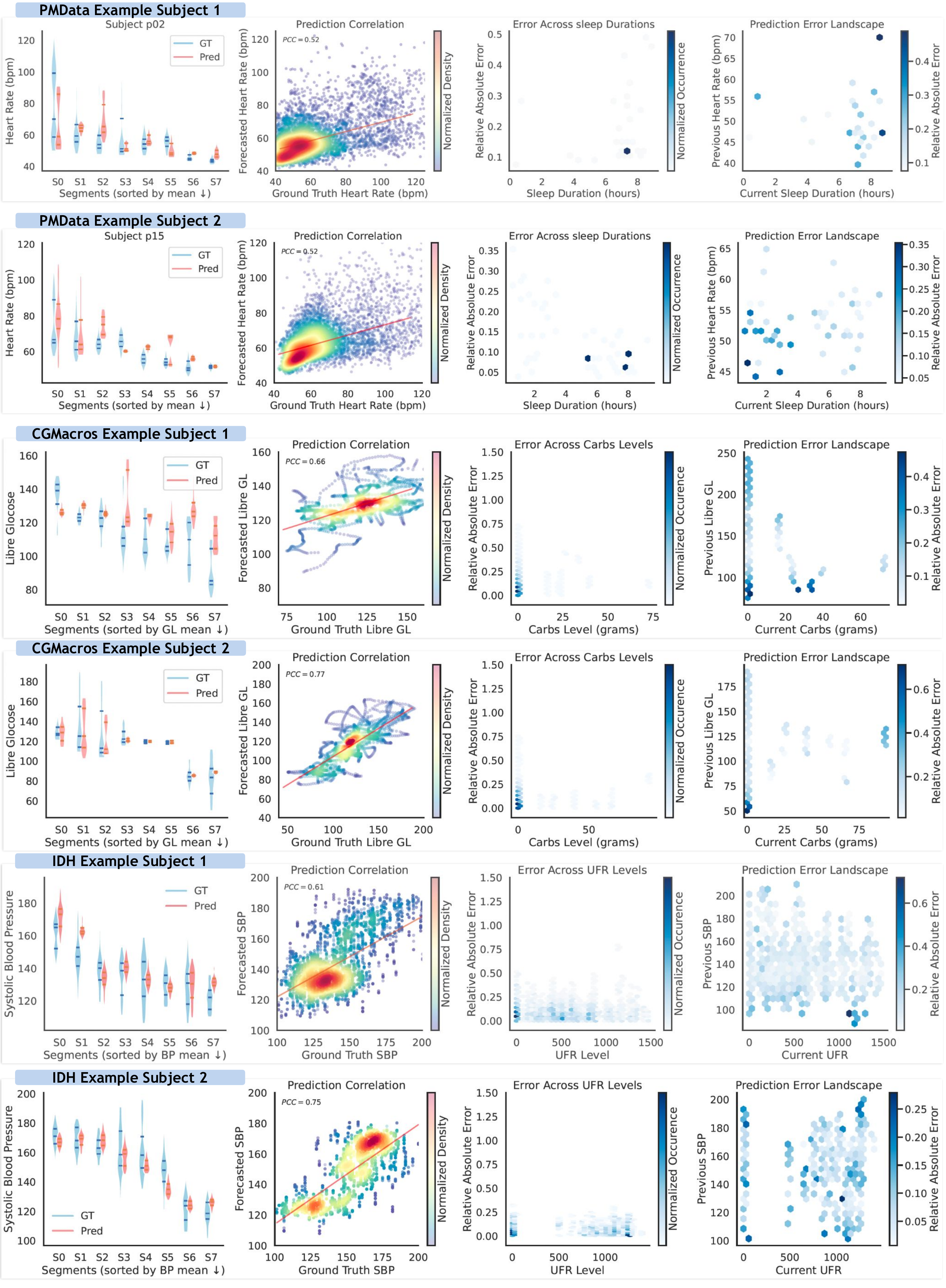}
\vspace{-3mm}
\caption{
Supplementary results for Figure \ref{fig:quantitative_results} panel D: Action/intervention analysis from example subjects.
}
\label{fig:additional_action_analysis}
\end{figure*}

\clearpage
\section{Procedure of inspect extent of balance using chaos theory based metrics} \label{app:nld-cluster-process}

The balance inspection involve 3 main stages: (I) metrics computation, (II) unsupervised clustering, and (III) balance score computation. 
The first stage generally follows the procedure described in algorithm \ref{alg:nld}, which is implemented with multi-process manner as demonstrated in the codebase. 
The second stage follows the entirely automated pipeline described in algorithm \ref{alg:clustering}, where the outcome is visualized in section \ref{app:chaotic-cluster}. 
Finally, the balance score is computed based on the outcome from the previous two stages, where the scoring details is presented in section \ref{app:data-balance}. 
In addition, the pseudo-code for offline balancing using the chaos metrics is presented in algorithm \ref{alg:balancing}. 

\begin{algorithm}[htbp!]
\caption{Clustering and Dynamical-System Typing Based on Chaos Metrics}
\label{alg:clustering}

\textbf{Input:} Chaos feature matrix $\mathbf{F} \in \mathbb{R}^{N \times M}$\\
\textbf{Step 1: Determine optimal cluster number.}

{\hangindent=1.5em
Compute K-means for $k = 2,3,\dots,K_{\max}$.\par}
{\hangindent=1.5em
Record final inertia value for each $k$.\par}
{\hangindent=1.5em
Select $k^\star \gg k_{\text{optimal}}$, where $k_{\text{optimal}}$ is identified using the elbow rule.\par}

\vspace{0.5em}
\textbf{Step 2: Clustering}

{\hangindent=1.5em
Fit K-means with $k^\star$ and obtain centroids $\{c_1, \dots, c_{k^\star}\}$.\par}

\vspace{0.5em}
\textbf{Step 3: Assign semantic labels to centroids.}

{\hangindent=1.5em
Compute global feature means $\mu$ across all centroids.\par}

\vspace{0.3em}
\textbf{for} $j = 1$ \textbf{to} $k^\star$ \textbf{do}

\hspace{2em}
{\hangindent=3em
Initialize label string $L_j$.\par}

\vspace{0.3em}
\hspace{2em}
{\hangindent=3em
\textit{// DFA (correlation or stationarity):}\par}

\hspace{2em}
{\hangindent=4.5em
\textbf{If} $c_j[\mathrm{DFA}] < 0.5$, append ``Anti-corr'' to $L_j$.\par}
\hspace{2em}
{\hangindent=4.5em
\textbf{Else if} $c_j[\mathrm{DFA}] < 1.0$, append ``Positive-corr'' to $L_j$.\par}
\hspace{2em}
{\hangindent=4.5em
\textbf{Else}, append ``Non-station'' to $L_j$.\par}

\vspace{0.3em}
\hspace{2em}
{\hangindent=3em
\textit{// Lyapunov exponent (degree of chaos):}\par}

\hspace{2em}
{\hangindent=4.5em
\textbf{If} $c_j[\lambda] < 0$, append ``Stable'' to $L_j$.\par}
\hspace{2em}
{\hangindent=4.5em
\textbf{Else if} $c_j[\lambda] < \mu[\lambda]$, append ``Rel Chaos'' to $L_j$.\par}
\hspace{2em}
{\hangindent=4.5em
\textbf{Else}, append ``Rel Very Chaos'' to $L_j$.\par}

\vspace{0.3em}
\hspace{2em}
{\hangindent=3em
\textit{// Persistent entropy (topological complexity):}\par}

\hspace{2em}
{\hangindent=4.5em
\textbf{If} $c_j[\mathrm{PE}_{H0}] < \mu[\mathrm{PE}_{H0}]$, append ``Low Connect Complex'' to $L_j$.\par}
\hspace{2em}
{\hangindent=4.5em
\textbf{Else}, append ``High Connect Complex'' to $L_j$.\par}
\hspace{2em}
{\hangindent=4.5em
\textbf{If} $c_j[\mathrm{PE}_{H1}] < \mu[\mathrm{PE}_{H1}]$, append ``Low Loop Complex'' to $L_j$.\par}
\hspace{2em}
{\hangindent=4.5em
\textbf{Else}, append ``High Loop Complex'' to $L_j$.\par}

\vspace{0.3em}
\hspace{2em}
{\hangindent=3em
Assign $L_j$ as the type of centroid $c_j$.\par}

\vspace{0.3em}
\textbf{end for}

\vspace{0.5em}
\textbf{Step 4: Merge clusters with identical type labels.}

{\hangindent=1.5em
Group centroids sharing the same label and compute histogram.\par}

\vspace{0.3em}
\textbf{Return:} merged cluster types and histogram

\end{algorithm}

\begin{algorithm}[ht!]
\caption{Compute Nonlinear Dynamics Metrics}
\label{alg:nld}

\textbf{Input:} Pretraining dataset $\mathcal{D}$ of multichannel time series\\
\textbf{Output:} NLD metrics per channel: DFA, Persistence Entropy, Lyapunov Exponent

\vspace{0.5em}
\textbf{for each} sample $X \in \mathcal{D}$ \textbf{do}

\hspace{1em}\textbf{for each} channel $x$ in $X$ \textbf{do}

\hspace{2em}\textbf{DFA:}\\
{\hangindent=2.5em
$d \leftarrow \mathrm{DFA}(x)$\par}

\vspace{0.4em}
\hspace{2em}\textbf{Persistence Entropy:}\\
{\hangindent=2.5em
$\tilde{x} \leftarrow \mathrm{TakensEmbed}(x;\ \text{delay}=1,\ \text{dim}=5)$\\
$D \leftarrow \mathrm{VietorisRipsPersistenceDiagram}(\tilde{x})$\\
$p \leftarrow \mathrm{PersistenceEntropy}(D)$\par}

\vspace{0.4em}
\hspace{2em}\textbf{Lyapunov Exponent:}\\
{\hangindent=2.5em
$\lambda \leftarrow \mathrm{LyapunovExponent}(x;\ \text{embdim}=10,\ \tau=1,\ \text{minsep}=10)$\par}

\vspace{0.4em}
\hspace{2em}\textbf{store} $(d, p, \lambda)$

\vspace{0.3em}
\hspace{1em}\textbf{end for}

\vspace{0.3em}
\textbf{end for}

\end{algorithm}

\begin{algorithm}[htbp!]
\caption{Iterative Chaos-Balance-Aware Sampling}
\label{alg:balancing}
\begin{algorithmic}[1]
\STATE \textbf{Input:} Dataset $\mathcal{D}$ consisting of $M$ data sources $\{\mathcal{D}_1, \dots, \mathcal{D}_M\}$

\STATE Shuffle samples within each data source $\mathcal{D}_m$
\STATE Initialize selected set $\mathcal{S} \leftarrow \emptyset$
\STATE Initialize remaining samples $\mathcal{R}_m \leftarrow \mathcal{D}_m$ for all $m$

\WHILE{$|\mathcal{S}| < 0.5 \times |\mathcal{D}|$}
    \FOR{$m = 1$ to $M$}
        \IF{$\mathcal{R}_m \neq \emptyset$}
            \STATE Randomly sample $x$ from $\mathcal{R}_m$
            \IF{$\mathrm{Score}(\mathcal{S} \cup \{x\}) > \mathrm{Score}(\mathcal{S})$}
                \STATE $\mathcal{S} \leftarrow \mathcal{S} \cup \{x\}$
            \ENDIF
            \STATE Remove $x$ from $\mathcal{R}_m$
        \ENDIF
    \ENDFOR
\ENDWHILE

\STATE \textbf{Return} $\mathcal{S}$
\end{algorithmic}
\end{algorithm}

\clearpage
\section{Clustering of Time Series Systems with Chaos Metrics} \label{app:chaotic-cluster}

\begin{figure*}[hb!]
  \centering
\makebox[\textwidth][c]{%
        \includegraphics[width=0.9\textwidth]{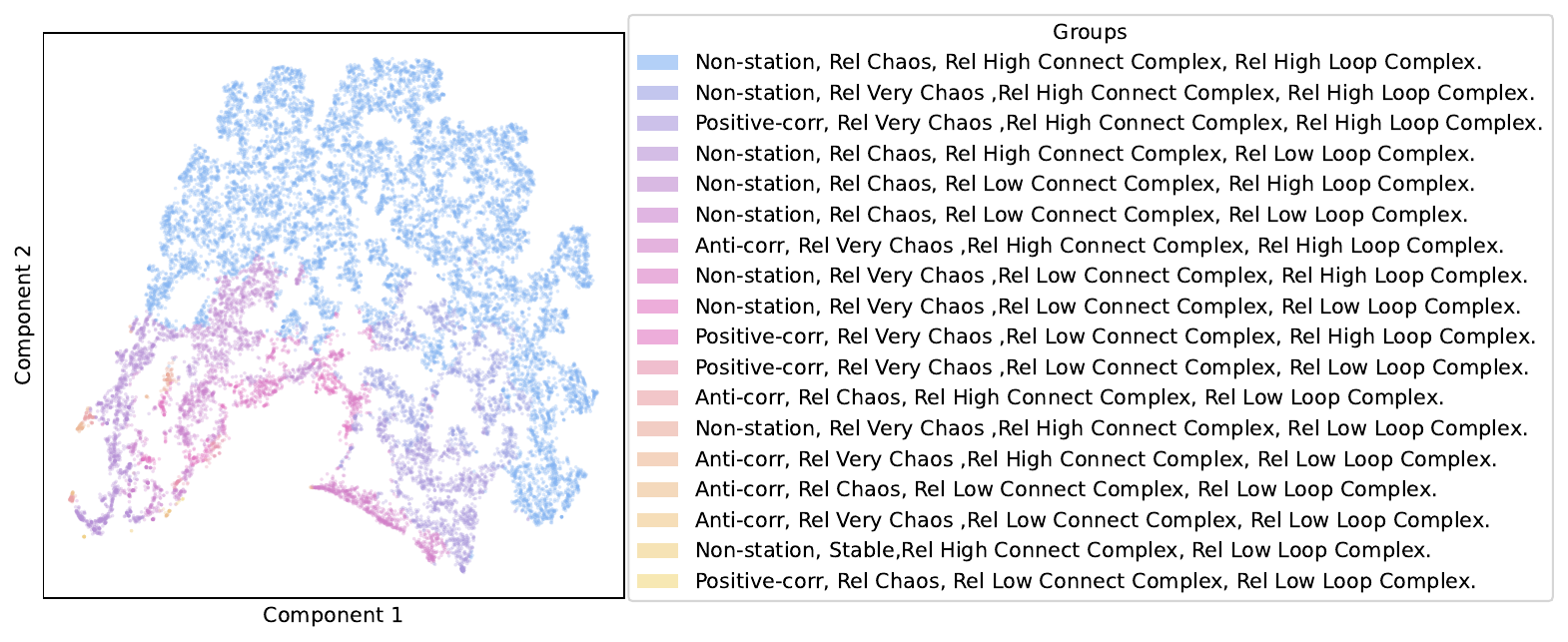}}
\makebox[\textwidth][c]{%
        \includegraphics[width=0.9\textwidth]{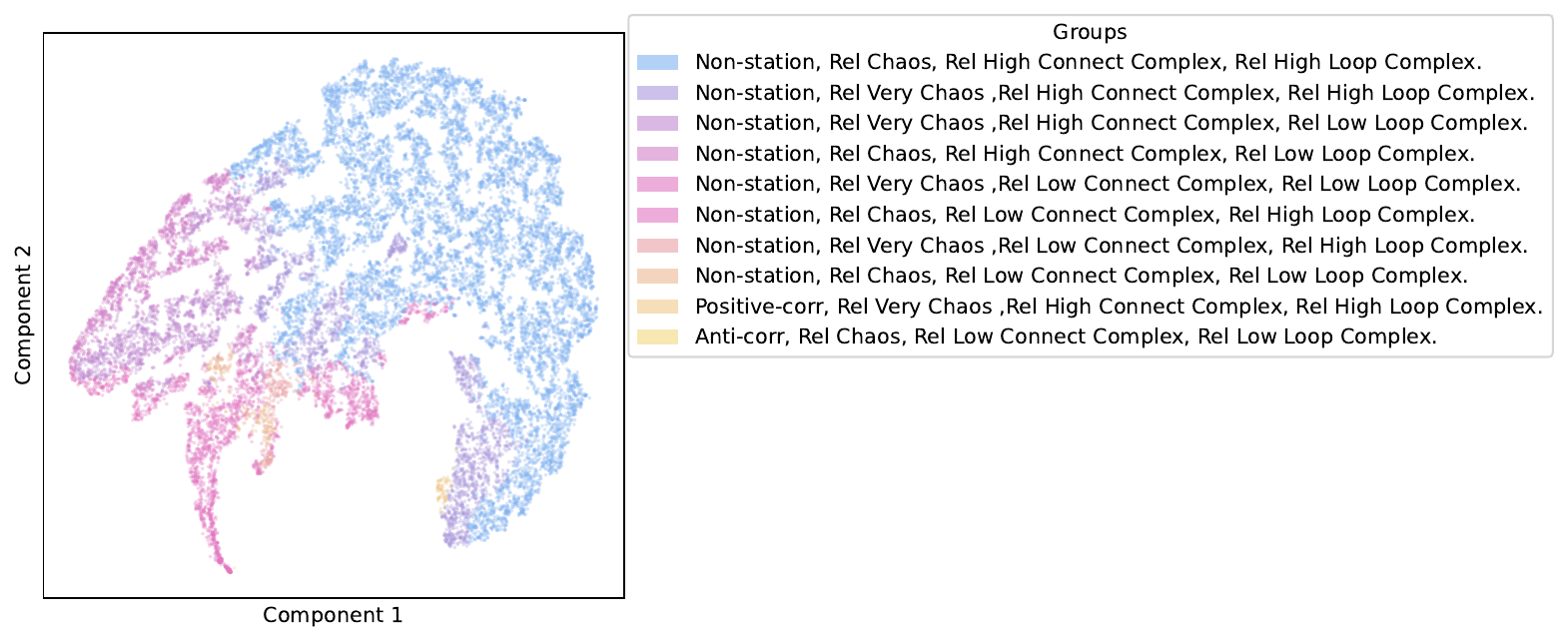}}
\makebox[\textwidth][c]{%
        \includegraphics[width=0.9\textwidth]{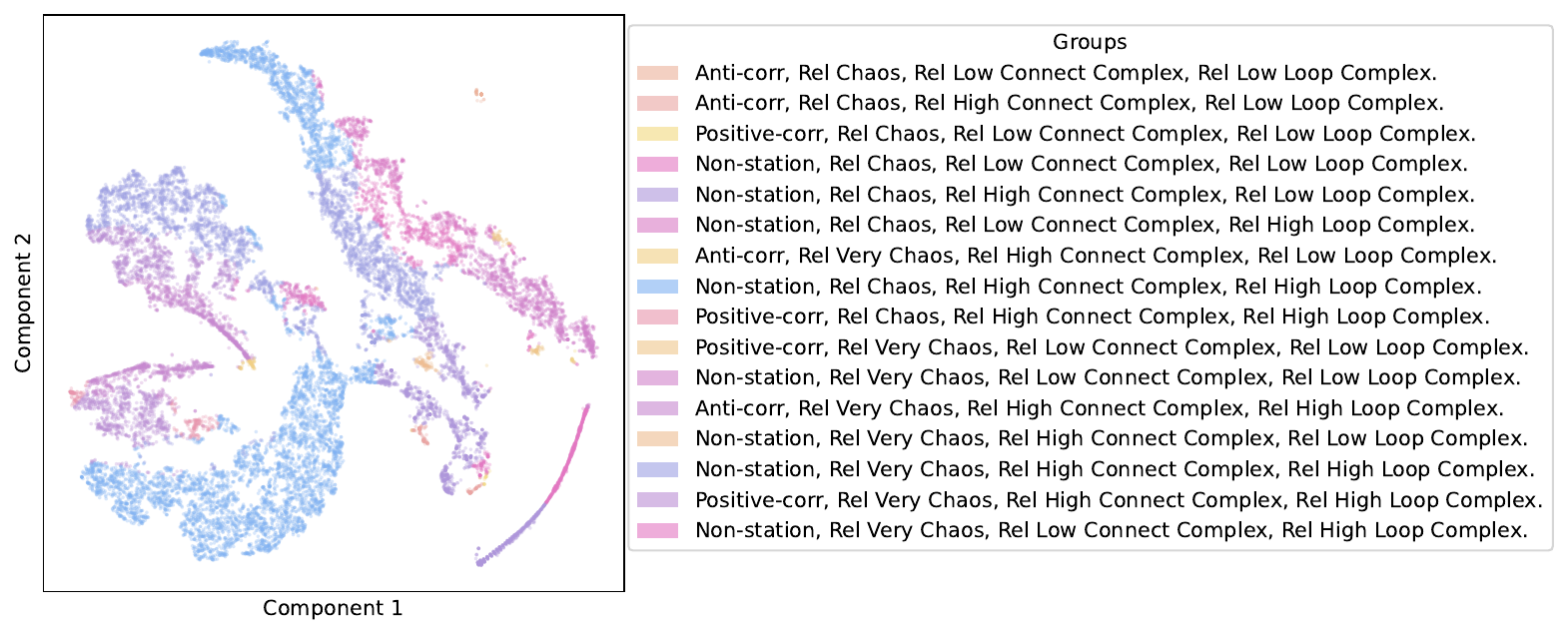}
        }
  \caption{\textbf{T-SNE plot of the datasets, quantified by the proposed metrics from chaotic theory}. This figure mainly specified the exact group of time series with different chaotic attributes corresponding to plot presented in 
  Figure \ref{fig:balance_results} panel A. 
  }
\label{fig:chaotic-cluster}
\end{figure*}

The details of the description of varied time series system in term of the chaotic theory based metrics are presented in Figure \ref{fig:chaotic-cluster}.

\clearpage

\section{\edit{Model Implementation Detail}}
\label{app:implement-detail}
\edit{
We provide a formal description of the masking, encoding, and reconstruction pipeline used in \textsc{NormWear-2}. Let the input multivariate time series be
$X \in \mathbb{R}^{C \times T}$
where $C$ denotes the number of channels and $T$ the sequence length.
\subsection{Patchification}
Each channel is divided into non-overlapping patches of length $P$:
\begin{equation}
    X^{(c)} \rightarrow 
    \big\{ x^{(c)}_1, x^{(c)}_2, \dots, x^{(c)}_L \big\},
    \qquad x^{(c)}_i \in \mathbb{R}^{P},
\end{equation}
where $L = T / P$. Each patch is projected to a $D$-dimensional embedding through a Conv1D patch embedding module $E(\cdot)$:
\begin{equation}
    z^{(c)}_i = E\big(x^{(c)}_i\big) \in \mathbb{R}^{D}.
\end{equation}
where $D$ is the latent dimension. 
\subsection{Channel-wise Mask Sampling}
Masks are then sampled independently for each channel and each patch:
\begin{equation}
    m^{(c)}_i \sim \mathrm{Bernoulli}(p_{\mathrm{mask}}),
\end{equation}
where $m^{(c)}_i$ is the result of the indicator function with success or failure outcome sampled from the Bernoulli distribution. The masked embedding is then formed as
\begin{equation}
    \tilde{z}^{(c)}_i = 
    \big(1 - m^{(c)}_i\big) \cdot z^{(c)}_i
    \; + \;
    m^{(c)}_i \cdot [\text{MASK}].
\end{equation}
where [MASK] is a single trainable vector similar to the [CLS] special token. This produces patch-level masked embeddings for all channels. Then these embeddings are sending to the core backbone as described in method section \ref{sec:backbone}. Denote the output from the backbone as $\hat{z}^{(c)}_{i, latent} \in \mathbb{R}^{D}$. 
\subsection{Patch-Level Deconvolution}
The latent embeddings generated from the backbone are then projected back to waveform patches through a lightweight DeConv1D module:
\begin{equation}
    \hat{s}^{(c)}_i = \mathrm{DeConv1D}\!\left(\hat{z}^{(c)}_{i, latent}\right)
    \in \mathbb{R}^{P \times K}
\end{equation}
Such a deconvolve step generates multiple candidate reconstructions $\{\hat{s}^{(c,k)}_i\}_{k=1}^K$ where $K$ is the maximum number of candidates. These candidate reconstructions are then consolidated via a Conv1D module:
\begin{equation}
    \hat{s}^{(c)}_i
    =
    \mathrm{Conv1D}
    \Big(
        \hat{s}^{(c,1)}_i,
        \dots,
        \hat{s}^{(c,K)}_i
    \Big) \in \mathbb{R}^{P \times 1}
\end{equation}
and collectively, the final reconstruction is:
\begin{equation}
    \hat{X_{\text{reconstruct}}} = \text{ConcatReshape}(\hat{s}^{(c)}_i , \;\;\forall c\in \{1, ..., C\}, i\in \{1, ..., L\}) \in \mathbb{R}^{C \times T}.
\end{equation}
}

\section{Model and Training Configuration} \label{app:train-config}
NormWear-2 is derived from the Masked Autoencoder (MAE) \citep{mae}. The detailed hyper-parameter choice is describe in \ref{tab:model-hyper}. We use a Conv1D layer with a kernel size of 16 and a stride of 16, ensuring no overlapping patches. This layer takes input with 1 channels and projects it to 768 channels, matching the hidden size of our encoders. In NormWear-2, we apply random masking independently to each variate along both the frequency and time axes, with respective masking ratios of 0.5. The masked patch are replaced by a trainable mask special token before passing to the encoder. To enhance representation learning, following \citet{normwear}, we introduce six additional transformer blocks as fusion layers, interleaved with the original 12 encoder blocks, creating a total of 18 blocks. Each transformer block has a hidden dimension of 768 and uses LayerNorm as in the original MAE. The latent embeddings obtained from the encoder are projected from 768 to 512 dimensions before passing to the decoding blocks. The positional embeddings are added to guide the decoder in reconstructing the input series. The lightweight decoder consists of two transformer blocks with a hidden dimension of 512, followed by two Conv1D layers. The first Conv1D layer maps from the flattened multivariate signal embedding to an intermediate dimension, and the second Conv1D layer maps from this intermediate dimension back to the original multivariate signal space. A GELU activation function is used between these layers, with BatchNorm applied to the input. The decoder reconstructs the original input series, and the model is trained using Mean Squared Error (MSE) loss on all data points. 

\begin{table}[ht]
    \centering
    \renewcommand{\arraystretch}{1.1} 
    \setlength{\tabcolsep}{4pt} 
        \centering
        \caption{\textbf{Pretraining Hyper-parameters.}}
        \begin{tabular}{l|r}
        \hlineB{2}
        \hlineB{2}
        \bf Hyper-parameter & \bf Value \\
        \hline
        \# cross-patches Transformer Encoder & 12 \\
        \# cross-channels Transformer Encoder & 6 \\
        \# Transformer Decoder & 2 \\
        \# Attention Heads & 12 \\
        Encoder Latent Size & 768 \\
        Decoder Latent Size & 512 \\
        Feedforward Latent Size & 3072 \\
        Normalization & LayerNorm \\
        Patch size & 16 \\
        \hlineB{2}
        Optimizer & AdamW \\
        Loss Scalar & NativeScaler \\
        Base Learning Rate (blr) & 5e-4 \\
        Epochs & 100 \\
        Batch size & 128 \\
        \hlineB{2}
        \hlineB{2}
        \end{tabular}
        \label{tab:model-hyper}
\end{table}

The models are pretrained on 8 NVIDIA RTX 3090 graphical computing unit (GPU), with 24GB of GPU memory on each card, along with 32 CPU cores, and 64GB of RAM. All the evaluation, analysis, visualization are conducted on a separate machine, with 1 NVIDIA RTX 4090 GPU, 24GB of GPU memory, 32 CPU cores, and 64GB of RAM. 

\section{Complexity Analysis of Varied Channel-Aware Encoding Mechanism} \label{app:channel-aware-complex}
Among the three core pre-trained model comprised in this study, they leverage similar idea but different implementation to approach the multivariate time series modeling. \textsc{Sundial} \citep{sundial} propose \textit{single-series sequence} (S3) for which they aggregate all the input series channels into a single channel along the temporal axis. When passing the time series in S3 format to the backbone model, the attention-based encoding schema is equivalent to the \textit{All-Attention} mechanism discussed in \citet{normwear}, which, as the authors stated, have the complexity of
\begin{equation}
    M_{Sundial} = O(d\cdot (L\cdot C)^2)
\end{equation}
where $d$ is the embedding size, $L$ as the sequence length, and $C$ is the number of input variate or series channel. Such running complexity from the S3 approach scaled up with the production of between sequence length and the number of input channels in a polynomial manner. In comparison, \textsc{NormWear}'s final optimal channel-aware encoding schema \citep{normwear}, which is also leveraged in this study, have the complexity of 
\begin{equation}
    M_{NormWear} = O(d\cdot C^2)
\end{equation}
which only scale up with the input number of variates. Lastly, \textsc{Panda} \citep{panda} proposed a very similar channel-aware mechanism, which the self-attention is applied on the input variate dimension across data representation from all time series. Such design has complexity equivalent to the \textit{Cross-Attention} as analyzed by \citet{normwear}:
\begin{equation}
    M_{Panda} = O(d\cdot L \cdot C^2)
\end{equation}
Finally, we can conclude that
\begin{equation}
    M_{Sundial} > M_{Panda} > M_{NormWear}
\end{equation}

\section{Evaluating Data Balance} \label{app:data-balance}
To evaluate the extent of data balance in terms of the metrics from chaotic theory as leveraged in this study, we mainly consider two main aspects, namely homogeneity and granularity of the distribution as presented in Figure \ref{fig:balance_results} section A. Varied approaches for inspecting the balance attributes are presented below. 

\subsection{Unnormalized Shannon Entropy.}
Shannon entropy \citep{shannon-entropy} is one of the metrics widely used to evaluate the amount of information within a probability distribution:
\begin{equation}
    H(p) = - \sum_{i=1}^n p_i \log (p_i)
\end{equation}
where $p$ denote a set of probability sum up to 1. Such entropy value not only reflect homogeneity of a distribution, but also comprise granularity information, which is indicated by the fact that the more group of system that can be clustered from a dataset, the more likely the higher the value of $H(p)$. 

\subsection{Weighted sum of normalized Shannon Entropy and Granularity.} \label{app:entropy_granular}
Since normalized Shannon entropy (with denominator of $\log(n)$) is often used under different scenario, we need to have a separate metric to explicitly evaluate the extent of granularity of a distribution. To achieve this, we leveraged a straightforward scoring formula that reflect the relative granularity:
\begin{equation}
    G(p) = \frac{|p|}{\max_{p' \in P} |p'|}
\end{equation}
where $P$ is the collection of all the distributions in comparison, and $|p|$ indicate the number of bins or possible outcome of a particular distribution. The final balance score $B(p)$ is then expressed as a weighted sum of:
\begin{equation}
    B(p) = \alpha \cdot \frac{H(p)}{\log(|p|)} + (1-\alpha) \cdot G(p)
\end{equation}
Since the value range of $H(p)$ and $G(p)$ is different, with relative ratio of total scores of around $4:6$, we use $\alpha = 0.6$ to balance this metric. 

\subsection{Weighted sum of Coefficient of Variation and Granularity.}
Another commonly used metric that also reflect the extent of homogeneity of a distribution is coefficient of variation \citep{coefficient-variation}, which is built on an assumption that the distribution is approximately a Gaussian distribution. From Figure \ref{fig:balance_results} panel A, we observed that most of the distribution is nearly a Gaussian distribution with different mean and variance. We then leverage this new metric similar to the metric in section \ref{app:entropy_granular}, with $\frac{H(p)}{\log(|p|)}$ replaced by $\frac{1}{CV(p)}$, where $CV(p)$ is defined by:
\begin{equation}
    CV(p) = \frac{\sigma_p}{\mu_p}
\end{equation}
Similarly, we use $\alpha = 0.5$ for the same reason in section \ref{app:entropy_granular} to balance the metric. 


\section{Detailed Data-Balance-Aware Study Evaluation Performance} \label{app:eval-balance-all}
Table \ref{tab:drop_drift} present the analysis result of the horizontal drift of models' performance. Table \ref{tab:result-overall} contains more fine-grained report of the performance of the methods in comparison. 

\begin{table}[htbp!]
\centering
\caption{
\edit{
Relative Drop in MAE (\%) and Average Drift across different methods between long term and short term generative series.
From the results, no consistent pattern is observed across models. For instance, Sundial shows the least horizontal drift between short-term and long-term predictions, but its absolute generative error remains higher than other approaches. Moreover, the drift speeds for all methods are on the order of $10^{-4}$, indicating that error accumulation over these horizons is minimal.
}
}
\label{tab:drop_drift}
\begin{adjustbox}{max width=1.0\linewidth}
\begin{tabular}{lccccc}
\toprule
\textbf{Metric} & \textbf{Sundial} & \textbf{Panda} & \textbf{NormWear-2} & 
\bf \begin{tabular}{@{\hspace{-0.5\tabcolsep}}l@{\hspace{-0.5\tabcolsep}}}
            \textsc{NormWear-2} \\
            Chaotic only
        \end{tabular} & 
\bf \begin{tabular}{@{\hspace{-0.5\tabcolsep}}l@{\hspace{-0.5\tabcolsep}}}
            \textsc{NormWear-2} \\
            Sensor only
        \end{tabular} \\
\midrule
Chaotic Drop      & +0.014\%  & -38.165\% & -13.208\% & -13.610\% & -2.162\% \\
Wearable Drop     & -3.560\%  & -14.821\% & -16.920\% & -15.690\% & -12.710\% \\
Civil Drop        & -19.955\% & -27.611\% & -24.994\% & -26.393\% & -23.223\% \\
Battery Drop      & -3.610\%  & +0.659\%  & -23.546\% & -17.491\% & -4.331\% \\
\textbf{Avg Drop} & -6.439\%  & -19.985\% & -19.667\% & -18.296\% & -10.607\% \\
\midrule
Chaotic Drift     & $4.7\times 10^{-5}$ & $6.6\times 10^{-4}$ & $3.4\times 10^{-4}$ & $3.5\times 10^{-4}$ & $7.6\times 10^{-5}$ \\
Wearable Drift    & $1.3\times 10^{-4}$ & $4.9\times 10^{-4}$ & $5.0\times 10^{-4}$ & $4.7\times 10^{-4}$ & $4.0\times 10^{-4}$ \\
Civil Drift       & $4.3\times 10^{-4}$ & $5.0\times 10^{-4}$ & $5.0\times 10^{-4}$ & $5.2\times 10^{-4}$ & $5.1\times 10^{-4}$ \\
Battery Drift     & $1.2\times 10^{-4}$ & $2.1\times 10^{-5}$ & $5.2\times 10^{-4}$ & $4.6\times 10^{-4}$ & $1.6\times 10^{-4}$ \\
\textbf{Avg Drift}& $1.6\times 10^{-4}$ & $4.1\times 10^{-4}$ & $4.6\times 10^{-4}$ & $4.5\times 10^{-4}$ & $2.9\times 10^{-4}$ \\
\bottomrule
\end{tabular}
\end{adjustbox}
\end{table}

\clearpage

\begin{table*}[ht!]
    \centering
    \caption{\textbf{Preliminary Evaluation Performance.} All the generative evaluation tasks are zero-shot forecasting on multivariate time series. The averaged results across test scenarios within each domain and each task are reported. For all the generative tasks, a lower MSE or MAE indicates a better prediction \citep{sundial}. For Battery SoH downstream tasks, mean absolute percentage error and $R^2$ are reported as the main metrics \citep{batterylife}. For the wearable downstream tasks, AUC-ROC is reported as the main metric \citep{normwear}. $1^{\text{st}}$ Count represents the number of wins achieved by a model across all domains and test scenarios. }
    \label{tab:result-overall}
    \resizebox{\textwidth}{!}{
    \renewcommand{\arraystretch}{1.0}
    \begin{tabular}{lccccc}
        \multicolumn{6}{c}{(a) \textbf{Generative Task Results}} \\
        \hlineB{2}
        \hlineB{2}
        \textbf{Evaluation Scenario} & 
        \textbf{\textsc{Sundial}}$_{large}$ & 
        \textbf{\textsc{Panda}} & 
          \bf \begin{tabular}{@{\hspace{-0.5\tabcolsep}}l@{\hspace{-0.5\tabcolsep}}}
            \textsc{NormWear-2} \\
            \;\;\;(Ours)
        \end{tabular}  & 
        \bf \begin{tabular}{@{\hspace{-0.5\tabcolsep}}l@{\hspace{-0.5\tabcolsep}}}
            \textsc{NormWear-2} \\
            Chaotic only
        \end{tabular}  & 
        \bf \begin{tabular}{@{\hspace{-0.5\tabcolsep}}l@{\hspace{-0.5\tabcolsep}}}
            \textsc{NormWear-2} \\
            Sensor only
        \end{tabular} \\
        \midrule
        \bf Metric & MAE $\downarrow$ \; MSE $\downarrow$ & MAE \; MSE & MAE \; MSE & MAE \; MSE & MAE \; MSE \\
        \hline
        Chaotic Short Forecast & 0.881 \; 1.222 & \bf 0.486	\; 0.587 & 0.678 \; 0.893 & \underline{0.652 \; 0.838} & 0.904 \; 1.342 \\
        Chaotic Long Forecast & 0.870 \; 1.175 & \bf 0.610 \; 0.785 & 0.770 \; 1.040 & \underline{0.750 \; 0.990} & 0.920 \; 1.380\\
        Chaotic Short Simulation & 0.874 \; 1.198 &	\bf 0.397 \; 0.464 &	\underline{0.629 \; 0.787} & 0.649 \; 0.839 & 0.900 \; 1.344 \\
        Chaotic Long Simulation & 0.860 \; 1.155 & \bf 0.610 \; 0.785 & 0.770 \; 1.040 & \underline{0.730 \; 0.925} & 0.935 \; 1.435\\
        \hline
        Chaotic Generative Avg. & 0.889 \; 1.241 & \bf 0.510 \; 0.638 & 0.645 \; 0.833 & \underline{0.631 \; 0.796} & 0.896 \; 1.319 \\
        \hlineB{2}
        Wearable Short Forecast & 0.930 \; 2.257 & 0.852 \; 2.236 & \bf 0.723 \; 1.555 & \underline{0.743 \; 1.631} &	0.765 \; 1.670 \\
        Wearable Long Forecast & 0.950 \; 2.170 & 1.010 \; 2.750 & \bf 0.880 \; 1.950 & \underline{0.880 \; 1.960} & 0.950 \; 2.330 \\
        Wearable Short Simulation & 0.924 \; 2.109 & 0.855 \; 1.885 & \bf 0.790 \; 1.789 & \underline{0.793 \; 1.821} & 0.840 \; 1.932 \\
        Wearable Long Simulation & 0.970 \; 2.260 & 0.950 \; 2.180 & \bf 0.900 \; 2.070 & \underline{0.900 \; 2.120} & 0.960 \; 2.300 \\
        \hline
        Wearable Generative Avg. & 0.953 \; 2.255 & 0.840 \; 1.933 & \bf 0.740 \; 1.538 & \underline{0.810 \; 1.813} & 0.835 \; 1.883 \\
        \hlineB{2}
        Civil Short Forecast & 0.539 \; 0.576 & \bf 0.413 \; 0.405 & 0.503 \; 0.561 & \underline{0.492 \; 0.535} & 0.533 \; 0.590 \\
        Civil Long Forecast & 0.654 \; 0.953 & \bf 0.557 \; 0.778 & 0.636 \; 0.938 & \underline{0.633 \; 0.925} & 0.678 \; 1.039 \\
        Civil Short Simulation & 0.569 \; 0.614 & \underline{0.509} \; \textbf{0.516} & 0.518 \; 0.569 & \textbf{0.508} \; \underline{0.549} & 0.582 \; 0.670 \\
        Civil Long Simulation & 0.675 \; 0.961 & \bf 0.619 \; 0.816 & 0.640 \; 0.925 & \underline{0.632 \; 0.897} & 0.697 \; 1.061 \\
        \hline
        Civil Generative Avg. & 0.609 \; 0.776 & \bf 0.525 \; 0.628 & 0.574 \; 0.748 & \underline{0.566 \; 0.727} & 0.622 \; 0.840 \\
        \hlineB{2}
        Battery Short Forecast & 0.926 \; 1.311 & \underline{0.616 \; 0.600} & \bf 0.512 \; 0.477 & 0.633 \; 0.695 & 0.951 \; 1.306 \\
        Battery Long Forecast & 0.947 \; 1.292 & \underline{0.702 \; 0.749} & \bf 0.671 \; 0.749 & 0.777 \; 0.954 & 1.005 \; 1.428 \\
        Battery Short Simulation & 0.791 \; 0.967 & 1.053 \; 1.628 & \bf 0.610 \; 0.642 & \underline{0.707 \; 0.770} & 0.977 \; 1.377 \\
        Battery Long Simulation & 0.833 \; 0.997 & 0.956 \; 1.333 & \bf 0.715 \; 0.759 & \underline{0.796 \; 0.914} & 1.007 \; 1.441 \\
        \hline
        Battery Generative Avg. & 0.874 \; 1.141 & 0.832 \; 1.077 & \bf 0.627 \; 0.657 & \underline{0.728 \; 0.833} & 0.985 \; 1.388 \\
        \hlineB{2}
        Short Forecast All. & 0.819 \; 1.341 & \textbf{0.592} \; 0.957 & \underline{0.604} \; \textbf{0.871} & 0.630 \; \underline{0.925} & 0.788 \; 1.227 \\
        Long Forecast All. & 0.855 \; 1.397 & \textbf{0.720} \; 1.265 & \underline{0.740} \; \textbf{1.174} & 0.762 \; \underline{1.216} & 0.875 \; 1.470 \\
        Short Simulate All. & 0.789 \; 1.222 & 0.704 \; 1.123 & \bf 0.637 \; 0.947 & \underline{0.664 \; 0.995} & 0.825 \; 1.331 \\
        Long Simulate All. & 0.834 \; 1.343 & 0.784 \; 1.278 & \bf 0.738 \; 1.149 & \underline{0.761 \; 1.192} & 0.884 \; 1.499 \\
        \hline
        Generative Micro Avg. & 0.780 \; 1.085 & \underline{0.668} \; 0.915 & \bf 0.638 \; 0.841 & 0.672 \; \underline{0.900} & 0.828 \;	1.234 \\
        \hlineB{2}
        \hlineB{2}\\
        \multicolumn{6}{c}{(b) \textbf{Downstream Task Results}} \\
        \hlineB{2}
        \hlineB{2}
        \bf Metric & MAPE $\downarrow$ \; $R^2$ $\uparrow$ & MAPE \; $R^2$ & MAPE \; $R^2$ & MAPE \; $R^2$ & MAPE \; $R^2$ \\
        \hline
        Battery Li-ion Downstream & \underline{0.154} \; 0.040 & 0.198 \; -0.726 & \bf 0.145 \; 0.229 & 0.202 \; -0.342 & 0.158 \; \underline{0.084} \\
        Battery Na-ion Downstream & \underline{0.050 \; 0.954} & 0.075 \; 0.879 & \bf 0.047 \; 0.963 & 0.070 \; 0.888 & 0.051 \; 0.947 \\
        Battery Zn-coin Downstream & 1.550 \; \underline{0.551} & 1.321 \; 0.255 & \bf 0.685 \; 0.563 & 0.964 \; 0.447 & \underline{0.695} \; 0.371 \\
        \hline
        Battery Downstream Micro Avg. & 0.628 \; 0.500 & 0.717 \; 0.073 & \underline{0.366} \; \textbf{0.531} & 0.396 \; 0.290 & \textbf{0.310} \; 0.328 \\
        \hlineB{2}
        \bf Metric & AUC ROC $\uparrow$ & AUC ROC & AUC ROC & AUC ROC & AUC ROC \\
        \hline
        Wearable State Recognition & 0.790 & 0.797 & \bf 0.813 & \underline{0.799} & 0.790 \\
        Wearable EEG Tasks & \bf 0.888 & 0.856 & \underline{0.872} & 0.862 & 0.864 \\
        Wearable Vital Sign (1-MAPE) & 0.948 & \bf 0.954 & \underline{0.953} & 0.949 & \underline{0.953} \\
        Wearable Disease Risk & 0.720 & 0.709 & \bf 0.759 & \underline{0.741} & 0.694 \\
        \hline
        Wearable Downstream Avg. & \underline{0.826} & 0.814 & \bf 0.838 & \underline{0.826} & 0.810 \\
        \hlineB{2}
        \bf $1^{\text{st}}$ Count & 12 & \underline{23} & \bf 30 & 5 & 4 \\
        \hlineB{2}
        \hlineB{2}
    \end{tabular}
    }
\end{table*}

\begin{table*}[ht!]
    \centering
    \caption{\textbf{Generative Results of Ablation Studies on Scaling in Pre-train Subsets with Varied Balance Scores.} }
    \label{tab:scaling-ablation-res}
    \begin{adjustbox}{max width=1.0\linewidth}
    \renewcommand{\arraystretch}{1.1}
    \begin{tabular}{l|ccc|ccc|ccc}
        \hlineB{2}
        \hlineB{2}
        Data Size & \multicolumn{3}{c|}{$10^3$} & \multicolumn{3}{c|}{$10^4$} & \multicolumn{3}{c}{$10^5$}\\\hline
        \textbf{Evaluation Scenario} & 
        \bf \begin{tabular}{@{\hspace{-0.5\tabcolsep}}l@{\hspace{-0.5\tabcolsep}}}
            Balance
        \end{tabular}  & 
        \bf \begin{tabular}{@{\hspace{-0.5\tabcolsep}}l@{\hspace{-0.5\tabcolsep}}}
            Chaotic
        \end{tabular}  & 
        \bf \begin{tabular}{@{\hspace{-0.5\tabcolsep}}l@{\hspace{-0.5\tabcolsep}}}
            Sensor
        \end{tabular} &
        \bf \begin{tabular}{@{\hspace{-0.5\tabcolsep}}l@{\hspace{-0.5\tabcolsep}}}
            Balance
        \end{tabular}  & 
        \bf \begin{tabular}{@{\hspace{-0.5\tabcolsep}}l@{\hspace{-0.5\tabcolsep}}}
            Chaotic
        \end{tabular}  & 
        \bf \begin{tabular}{@{\hspace{-0.5\tabcolsep}}l@{\hspace{-0.5\tabcolsep}}}
            Sensor
        \end{tabular} &
        \bf \begin{tabular}{@{\hspace{-0.5\tabcolsep}}l@{\hspace{-0.5\tabcolsep}}}
            Balance
        \end{tabular}  & 
        \bf \begin{tabular}{@{\hspace{-0.5\tabcolsep}}l@{\hspace{-0.5\tabcolsep}}}
            Chaotic
        \end{tabular}  & 
        \bf \begin{tabular}{@{\hspace{-0.5\tabcolsep}}l@{\hspace{-0.5\tabcolsep}}}
            Sensor
        \end{tabular}\\
        \midrule
        \bf Metric & MAE \; MSE & MAE \; MSE & MAE \; MSE & MAE \; MSE & MAE \; MSE & MAE \; MSE & MAE \; MSE & MAE \; MSE & MAE \; MSE\\
        \hline
        Chaotic Generative Avg. & 0.830 \; 1.064 & 0.803 \; 1.031 & 1.051 \; 1.791 & 0.698 \;	0.892 &	0.746 \; 0.962 & 1.079 \; 1.891 & 0.696 \; 0.900 & 0.695 \; 0.898 & 0.912 \; 1.369 \\
        Wearable Generative Avg. & 0.835 \; 1.829 &	0.841 \; 1.822 &	0.960 \; 2.218 & 0.826 \; 1.815	& 0.827 \; 1.809 & 0.935	\; 2.195 & 0.821 \; 1.836 &	0.828 \; 1.870 & 0.854 \; 1.930 \\
        Civil Generative Avg. & 0.608 \; 0.786 & 0.595 \; 0.762 & 0.731 \; 1.125 & 0.565 \; 0.717 & 0.575 \; 0.738 & 0.715 \; 1.105 & 0.574 \; 0.748 &	0.566 \; 0.727 & 0.622 \; 0.840 \\
        Battery Generative Avg. & 0.911 \; 1.075 & 0.875 \; 1.007 & 0.980 \; 1.337 & 0.846 \;	0.975 &	0.833 \; 0.960 & 1.028 \; 1.462 & 0.627 \; 0.657 & 0.728 \; 0.833 & 0.985 \; 1.388 \\
        \hlineB{2}
        Short Forecast All. & 0.752 \; 1.067 & 0.740 \; 1.050 & 0.910 \; 1.557 & 0.674 \; 0.933 & 0.690 \; 0.969 & 0.898 \; 1.505 & 0.604 \; 0.871 &	0.630 \; 0.925 & 0.788 \; 1.227 \\
        Long Forecast All. & 0.836 \; 1.324 & 0.817 \; 1.262 & 0.956 \; 1.705 & 0.771 \; 1.191	& 0.787 \; 1.216 & 0.959 \; 1.717 & 0.740 \;	1.174 &	0.762 \; 1.216 & 0.875 \; 1.470 \\
        Short Simulate All. & 0.770 \; 1.105 & 0.751 \; 1.069 &	0.912 \; 1.536 & 0.705 \; 1.018 & 0.710 \; 1.016 & 0.924 \; 1.596 & 0.637 \;	0.947 &	0.664 \; 0.995 & 0.825 \; 1.331 \\
        Long Simulate All. & 0.827 \; 1.259 & 0.806 \; 1.241 &	0.943 \; 1.673 & 0.785 \; 1.258 & 0.793 \; 1.268 & 0.976 \; 1.834 & 0.738 \;	1.149 &	0.761 \; 1.192 & 0.884 \; 1.499 \\
        \hline
        Generative Micro Avg. & 0.779 \; 1.036 & 0.757 \; 0.996 & 0.900 \; 1.422 & 0.715 \; 0.943 & 0.723 \; 0.957 & 0.915 \; 1.477 & 0.638 \;	0.841 &	0.672 \; 0.900 & 0.828 \;	1.234 \\
        \hlineB{2}
        \hlineB{2}
    \end{tabular}
    \end{adjustbox}
\end{table*}

\begin{table*}[t!]
\centering
\caption{\edit{Ablation Study: Performance comparison across models pretrained using different backbone architectures.}}
\resizebox{\textwidth}{!}{
\setlength{\tabcolsep}{4pt}
\renewcommand{\arraystretch}{1.0}
\begin{tabular}{lcccc}
\toprule
\textbf{Models} & 
\begin{tabular}{@{\hspace{-0.5\tabcolsep}}l@{\hspace{-0.5\tabcolsep}}}
            \textbf{Chronos$_{base}$}  \\
            \citep{chronos}
        \end{tabular} &
\begin{tabular}{@{\hspace{-0.5\tabcolsep}}l@{\hspace{-0.5\tabcolsep}}}
            \textbf{Uni-variate} \\
            \citep{patchtst}
        \end{tabular} & 
\begin{tabular}{@{\hspace{-0.5\tabcolsep}}l@{\hspace{-0.5\tabcolsep}}}
            \textbf{[CLS] Attn.}\\
            \citep{normwear}
        \end{tabular}  & 
\begin{tabular}{@{\hspace{-0.5\tabcolsep}}l@{\hspace{-0.5\tabcolsep}}}
            \textbf{Channel Attn.}\\
            \citep{panda}
        \end{tabular} \\
\midrule
\textit{Zero-shot Generative Tasks} & MAE $\downarrow$ & MAE $\downarrow$ & MAE $\downarrow$ & MAE $\downarrow$ \\
Short-term forecast & \textbf{0.433$\pm$0.018} & 0.564$\pm$0.019 & \underline{0.558$\pm$0.010} & 0.653$\pm$0.032 \\
Long-term forecast & \bf 0.638$\pm$0.083 & \underline{0.694$\pm$0.022} & \underline{0.696$\pm$0.019} & 0.747$\pm$0.030 \\
Short-term simulate & \underline{0.451$\pm$0.018} & 0.569$\pm$0.012 & 0.596$\pm$0.011 & \textbf{0.405$\pm$0.053} \\
Long-term simulate & \underline{0.696$\pm$0.135} & 0.706$\pm$0.019 & 0.702$\pm$0.018 & \textbf{0.477$\pm$0.062} \\
All short-term generative & \bf 0.442$\pm$0.017 & 0.566$\pm$0.015 & 0.577$\pm$0.010 & \underline{0.529$\pm$0.057} \\
All long-term generative & \underline{0.667$\pm$0.105} & 0.700$\pm$0.020 & 0.699$\pm$0.017 & \textbf{0.612$\pm$0.063} \\
All generative series & \bf 0.554$\pm$0.072 & 0.633$\pm$0.021 & 0.638$\pm$0.017 & \underline{0.571$\pm$0.060} \\
\midrule
\textit{Battery SOH Downstream} & $R^2 \uparrow$ & $R^2 \uparrow$ & $R^2 \uparrow$ & $R^2 \uparrow$ \\
Li-ion SOH & \underline{0.036$\pm$0.026} & -0.190$\pm$0.059 & \textbf{0.229$\pm$0.021} & -0.487$\pm$0.075 \\
Na-ion SOH & \textbf{0.966$\pm$0.000} & 0.525$\pm$0.255 & \underline{0.963$\pm$0.0002} & -1.628$\pm$3.145 \\
Zn-coin SOH & 0.433$\pm$0.017 & \bf 0.630$\pm$0.016 & \underline{0.563$\pm$0.012} & 0.541$\pm$0.054 \\
Battery SOH Avg & \underline{0.507$\pm$0.173} & 0.297$\pm$0.189 & \textbf{0.531$\pm$0.238} & -0.955$\pm$4.986 \\
\midrule
\textit{Wearable Downstream Tasks} & AUC-ROC $\uparrow$ & AUC-ROC$\uparrow$ & AUC-ROC $\uparrow$ & AUC-ROC $\uparrow$ \\
State Recognition & 0.799$\pm$0.11 & 0.804$\pm$0.025 & \underline{0.813$\pm$0.021} & \textbf{0.814$\pm$0.023} \\
EEG Tasks & 0.807$\pm$0.021 & \bf 0.875$\pm$0.027 & \underline{0.872$\pm$0.027} & \underline{0.872$\pm$0.026} \\
Vital Sign (1-MAPE) & \bf 0.953$\pm$0.001 & \underline{0.951$\pm$0.001} & \bf \textbf{0.953$\pm$0.001} & 0.949$\pm$0.001 \\
Disease Risk & 0.621$\pm$0.034 & \underline{0.743$\pm$0.037} & \textbf{0.759$\pm$0.034} & 0.724$\pm$0.045 \\
Wearable Avg. & 0.768$\pm$0.032 & \underline{0.832$\pm$0.028} & \textbf{0.838$\pm$0.025} & 0.826$\pm$0.031 \\
\hline
\bf Avg. Rank & \underline{2.250} & 2.688 & \bf 2.188 & 2.750 \\
\bottomrule
\end{tabular}
}
\label{tab:backbone_mae}
\end{table*}

\clearpage
\section{Qualitative Visualization} \label{app:qualitative-visual}
\begin{figure*}[ht!]
  \centering
  \makebox[\textwidth][c]{%
        \includegraphics[width=1.0\textwidth]{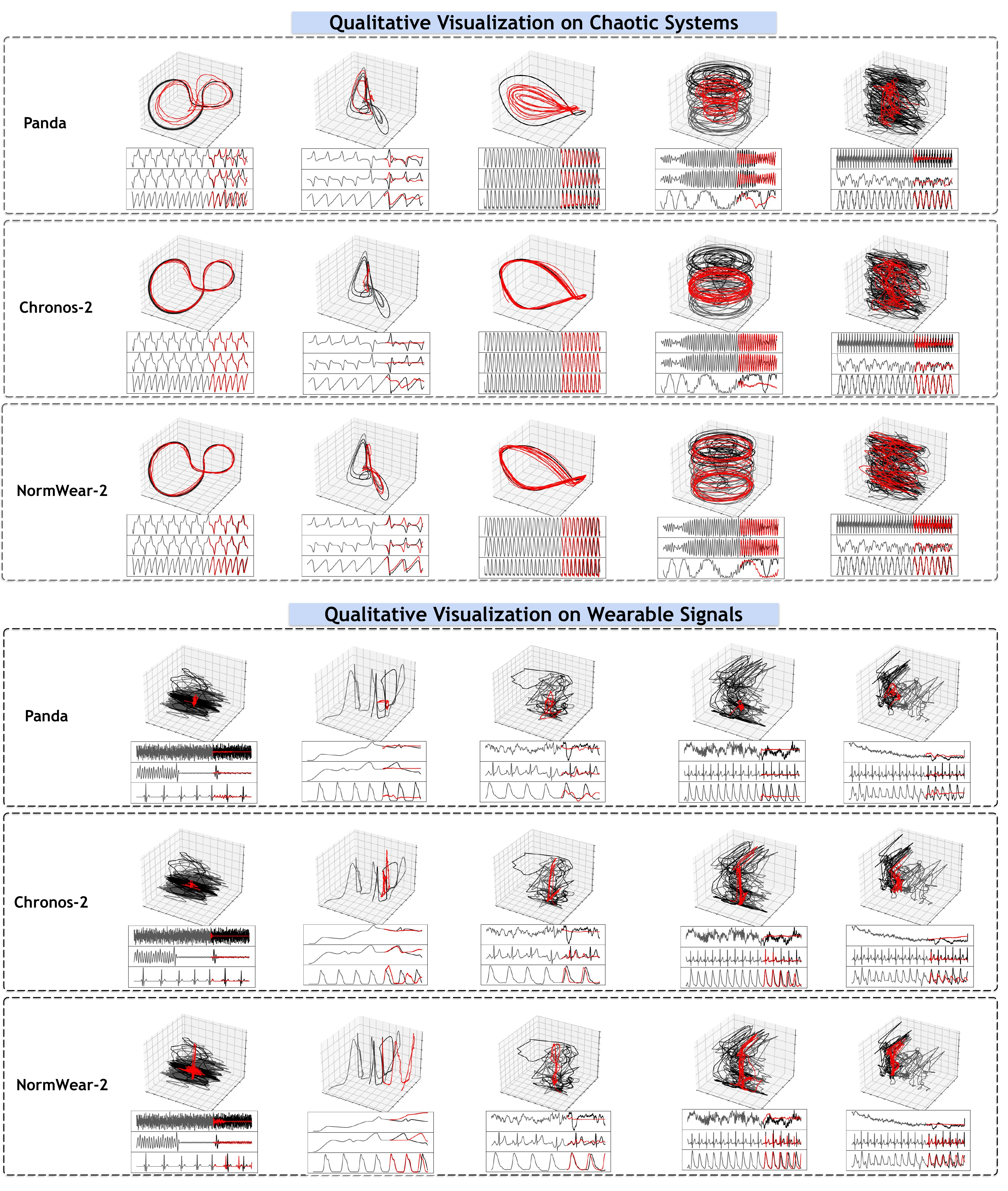}}
  \caption{\textbf{Visualization of generation on time series randomly generated from test set}. Models in comparison are Panda \citep{panda} and Chronos \citep{chronos}.}
\label{fig:quant-visual-compare}
\end{figure*}

\begin{figure*}[ht!]
  \centering
  \makebox[\textwidth][c]{%
        \includegraphics[width=1.0\textwidth]{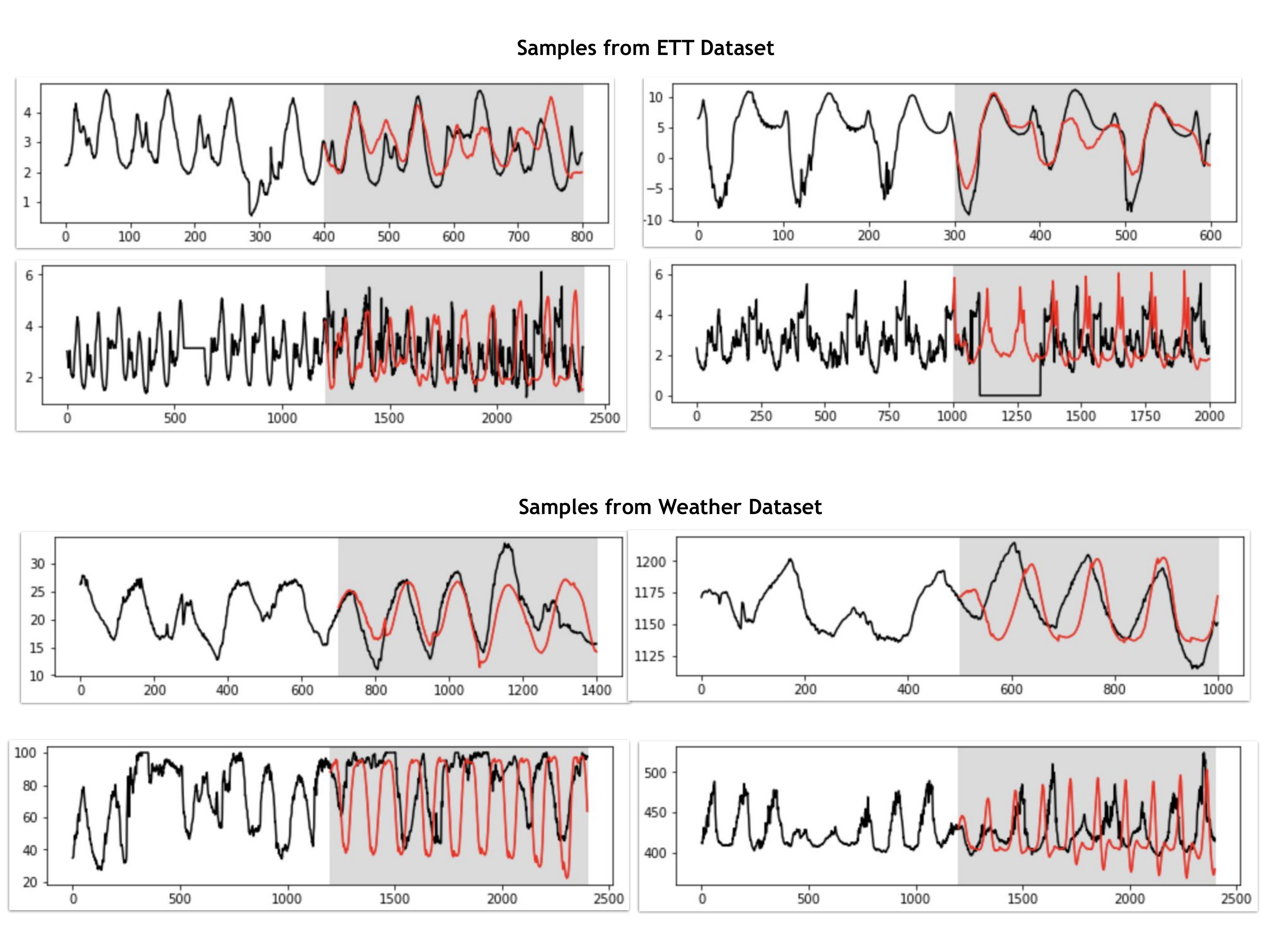}}
  \caption{\textbf{Visualization of generation on time series randomly generated from civil monitoring datasets proposed by \citet{autoformer}}.}
\label{fig:quant-visual-civil}
\end{figure*}

\begin{figure*}[hbtp!]
  \centering
  \makebox[\textwidth][c]{%
        \includegraphics[width=0.95\textwidth]{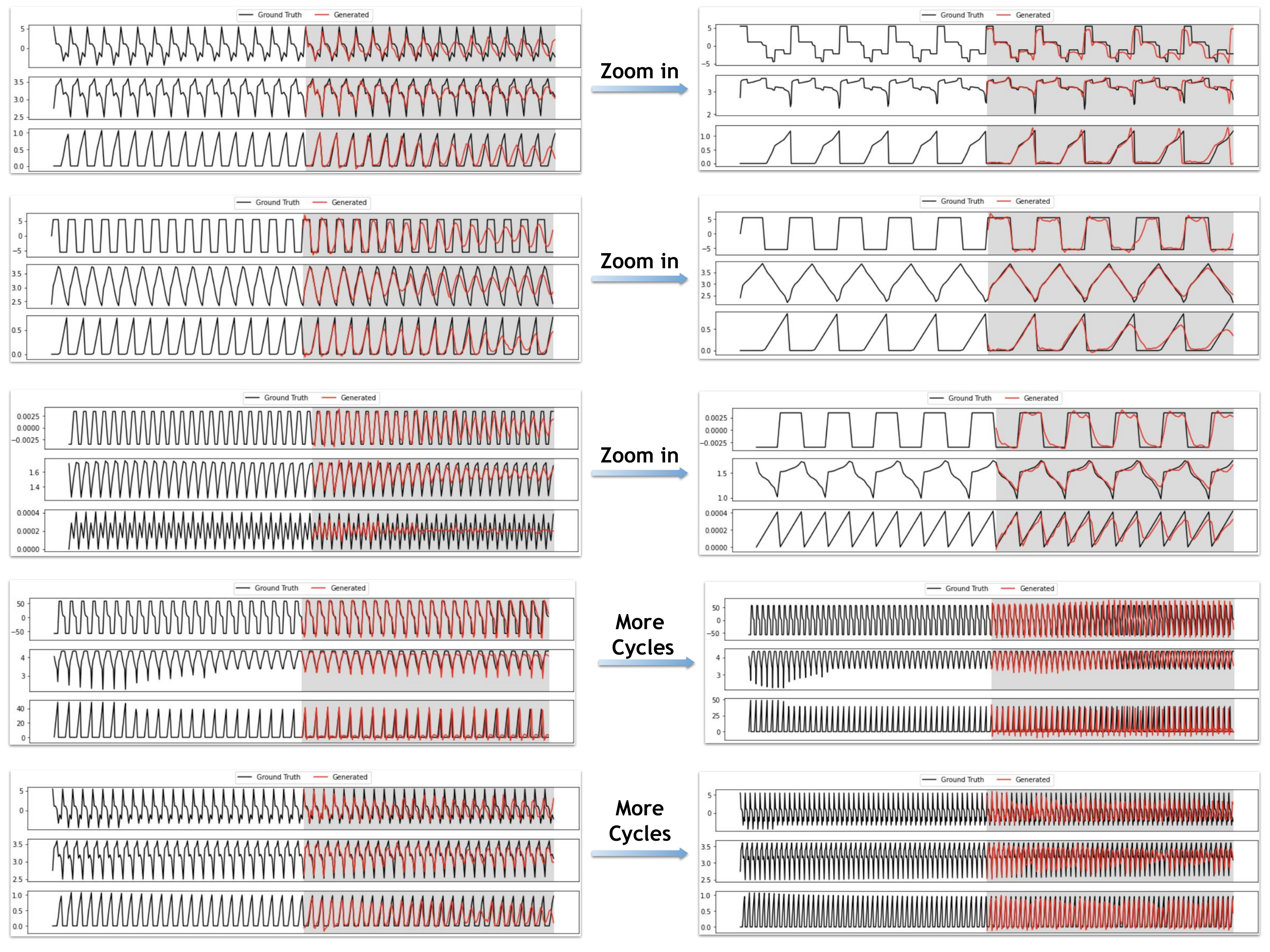}}
  \caption{\textbf{Visualization of generation on time series randomly generated from battery datasets proposed by \citet{batterylife}}.}
\label{fig:quant-visual-battery}
\end{figure*}

\begin{figure*}[hbtp!]
  \centering
  \makebox[\textwidth][c]{%
        \includegraphics[width=0.99\textwidth]{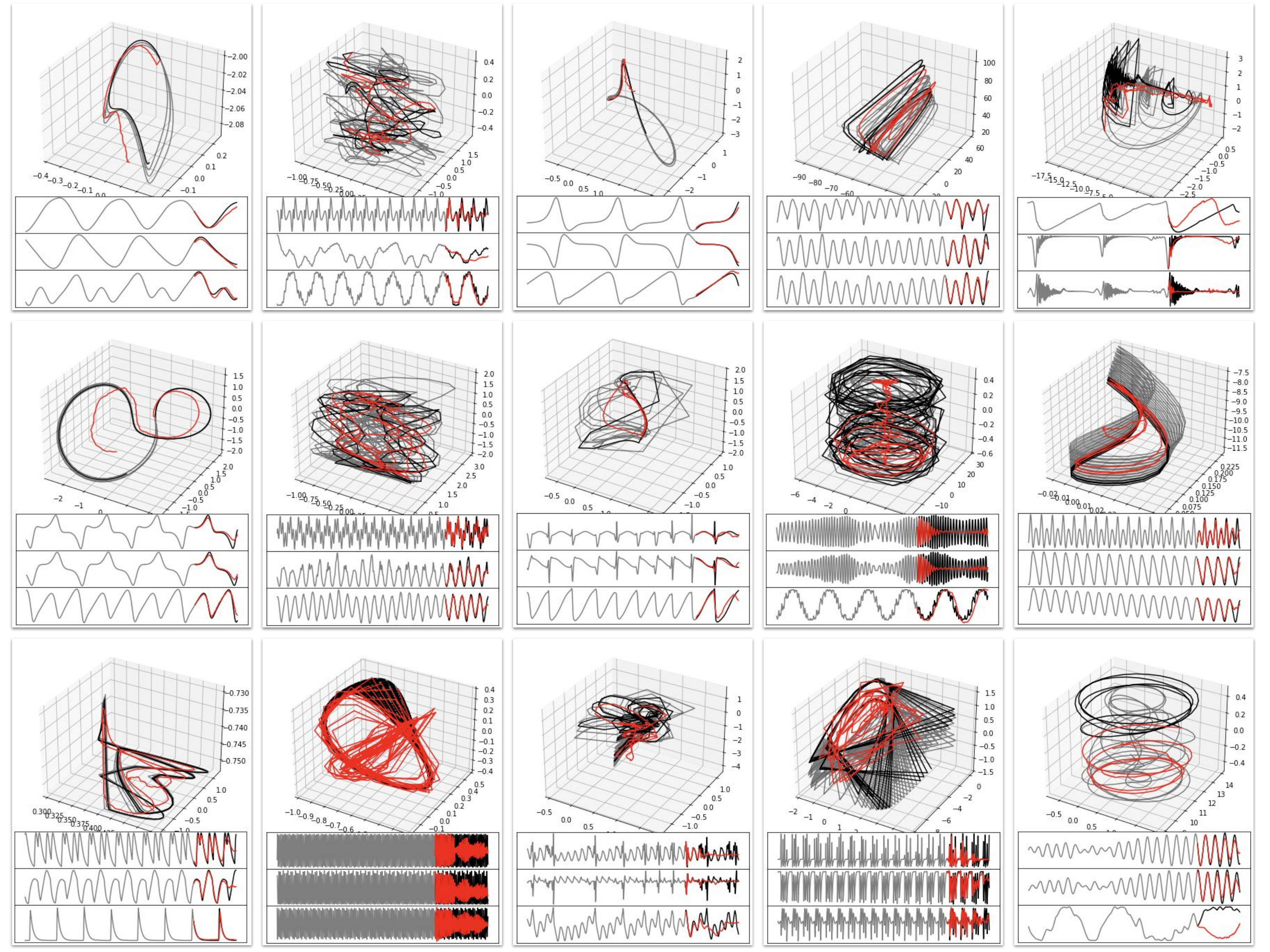}}
  \caption{\textbf{Visualization of generation on time series randomly generated from chaotic system datasets proposed by \citet{chaosbenchmark}}.}
\label{fig:quant-visual-chaotic}
\end{figure*}

Visualization of models' test time output is presented in Figure \ref{fig:quant-visual-compare}, \ref{fig:quant-visual-civil}, \ref{fig:quant-visual-battery}, and \ref{fig:quant-visual-chaotic}. 

\clearpage


\end{document}